\documentclass{article}

\PassOptionsToPackage{numbers, compress}{natbib}
 
 \usepackage[preprint]{style}

\usepackage[utf8]{inputenc} 
\usepackage[T1]{fontenc}    
\usepackage{hyperref}       
\usepackage{url}            
\usepackage{booktabs}       
\usepackage{amsfonts}       
\usepackage{nicefrac}       
\usepackage{microtype}      
\usepackage{xcolor}         

\usepackage{graphicx}
\usepackage{amsmath}
\usepackage{utfsym}  
\usepackage{multirow,algorithm}
\usepackage{colortbl}
\usepackage{enumitem}
\usepackage{amssymb}
\usepackage{makecell}
\usepackage{wrapfig}
\usepackage{bm}
\usepackage{tikz}
\usepackage{threeparttablex}
\usepackage{tabularx}

\usepackage{titletoc}

\newcommand{\todo}{\textcolor{black}}

\makeatletter
\renewcommand{\thanks}[1]{%
  \stepcounter{footnote}%
  \protected@xdef\@thanks{\@thanks\protect\footnotetext[\the\c@footnote]{#1}}%
}
\renewcommand{\@notice}{}
\makeatother

\title{
From Articulated Kinematics to Routed Visual Control for Action-Conditioned Surgical Video Generation
}

\author{ \\ 
  \small{Bohan Li$^{1,2,4*}$} \quad \small{Shuojue Yang$^{2*}$} \quad \small{Baorui Peng$^{4*}$} \quad \small{Xianda Guo$^{5*}$} \quad \small{Erli Zhang$^2$} \quad 
    \small{Youqi Tao$^6$} \quad 
  \\ \small{Junfeng Duan$^9$} \quad \small{Daguang Xu$^7$} \quad  \small{Qi Dou$^8$} \quad \small{Xin Jin$^4$} \quad \small{Wenjun Zeng$^4$} \quad  \small{Hao Zhao$^{3\dagger}$} \quad  \small{Yueming Jin$^{2\dagger}$}
   \\  \\ 
   \small{$^1$SJTU} \quad \small{$^2$NUS} \quad \small{$^3$THU} \quad \small{$^4$EIT} \quad \small{$^5$WHU} \quad \small{$^6$Harvard} \quad \small{$^7$NVIDIA} \quad \small{$^8$CUHK} \\ \quad \small{$^9$Shenzhen Hospital of Southern Medical University}
   \\ \small{Project page: \url{https://arlo0o.github.io/KVLR-project/}} \thanks{Contributed equally} \thanks{Corresponding} }

\begin{document}

\maketitle

\begin{figure*}[!ht]
\centering
\includegraphics[width=0.99\linewidth]{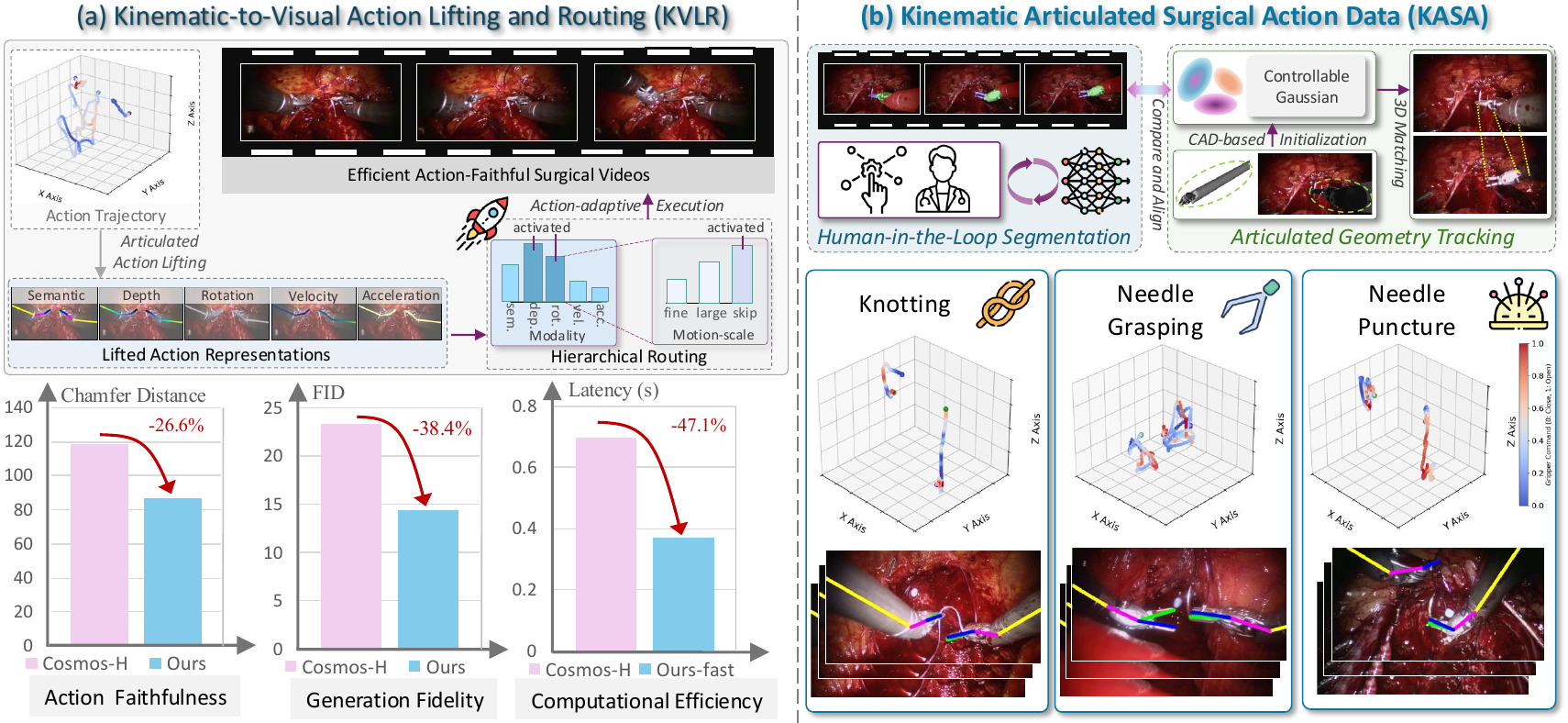}
\caption{
(a) KVLR transforms low-dimensional articulated kinematics into five image-aligned control modalities. A hierarchical routing mechanism selectively activates the most relevant modalities and motion scales, improving action faithfulness, visual fidelity, and computational efficiency.
(b) KASA provides articulated action annotations curated from real robotic surgical videos.\looseness=-1
}
 \vspace{-0pt}
\label{fig_teaser}
\end{figure*}

\begin{abstract}
Action-conditioned surgical video generation is a critical yet highly challenging problem for robotic surgery. The core difficulty is that low-dimensional control vectors must precisely govern complex image-space evolution. In this work, we propose a \textbf{kinematic-to-visual} lifting paradigm that converts articulated kinematics into a unified set of five image-aligned control modalities. Building on this representation, we introduce a \textbf{hierarchically routed visual control} framework that selectively activates the most relevant control modalities and motion scales. Instead of uniformly applying all control signals, our model performs hierarchical routing to dynamically allocate conditioning capacity. We further design kinematic-prior-guided routing loss functions to ensure physically meaningful, temporally stable, and efficient expert utilization. \textbf{To improve efficiency}, we propose a budgeted training and inference scheme that leverages routing-induced sparsity. By selectively discarding low-significance control pathways during training and execution, our approach enables adaptive computation that is complementary to standard distillation. We additionally construct \textbf{a new benchmark} with curated articulated annotations, obtained through human-in-the-loop semantic labeling and differentiable pose tracking, providing realistic supervision for action-conditioned surgical video generation. \textbf{Extensive experiments} demonstrate that our method consistently improves action faithfulness, visual fidelity, and cross-domain generalization over diverse baselines. Moreover, our efficient variant achieves substantial reductions in latency while maintaining strong control accuracy. 
\end{abstract}

\section{Introduction}

Robotic surgery is entering an era where perception and control must be tightly integrated, making action-conditioned surgical video generation a promising enabling tool~\cite{cho2024surgen,li2025ophora,chen2025surgsora,biagini2025hierasurg,wang2025medgen,ma2025open,turkcan2025towards}. Given a desired surgical action sequence, the ability to synthesize realistic and physically consistent visual outcomes has broad applications~\cite{biagini2025hierasurg,rapuri2026saw,he2026cosmosh}. For instance, surgeons or learning systems can simulate “what-if” scenarios (see Fig.~\ref{fig:generalization}) under different manipulation strategies. However, as illustrated in Fig.~\ref{fig_teaser} (a) bottom, this problem is fundamentally challenging: it requires simultaneously achieving high visual realism and precise action faithfulness, where low-dimensional articulated control signals must govern complex, high-dimensional image-space evolution.

The core difficulty lies in how to effectively introduce control signals into video generation~\cite{wan2025,kong2024hunyuanvideo,zheng2024opensora}. Text-based conditioning~\cite{cho2024surgen,he2026cosmosh,li2025ophora}, while flexible, is insufficient for surgical settings due to its ambiguity and lack of geometric grounding. Prior efforts have explored dense visual control signals (\textit{e.g.}, depth, segmentation, or flow)~\cite{chen2025surgsora,biagini2025hierasurg,rapuri2026saw,wang2025medgen}, which improve controllability but rely on expensive annotations or unavailable inputs in real robotic systems. This motivates the need for a unified representation that is both physically grounded and visually aligned.

We propose KVLR, a \textbf{K}inematic-to-\textbf{V}isual \textbf{L}ifting and \textbf{R}outing method (Sec.~\ref{sec:kva_field}) that transforms articulated kinematics into visual control. Specifically, we lift low-dimensional articulated kinematics into a pixel-aligned representation consisting of five interpretable modalities (semantics, depth, rotation, velocity, and acceleration), forming a unified visual control basis (Fig.~\ref{fig_teaser} (a) top). This kinematic-to-visual action field enables the model to explicitly encode where the tool is, how it is oriented, and how it moves. Compared to prior works~\cite{chen2025surgsora,rapuri2026saw,he2026cosmosh}, this representation is lightweight and 
can be directly derived from articulated actions commonly available in robotic settings.\looseness=-1

Building on this representation, KVLR allows a hierarchical routing mechanism (Sec.~\ref{sec:routing}) that dynamically allocates control capacity. As illustrated in Fig.~\ref{fig_teaser} (a) top, instead of uniformly applying all control modalities~\cite{zhang2023adding,yang2025orv,gao2026pam,zhao2023uni,lin2024ctrladapter}, our model performs modality-level routing to select the most relevant control modalities, and further applies motion-scale routing to specialize computation for fine manipulation, large transport motion, or static regions. To ensure that routing reflects physical structure rather than arbitrary patterns, we design kinematic-prior-guided loss functions (Sec.~\ref{sec:routing_learning}) that enforce physically meaningful expert utilization and temporal consistency.

An important product of hierarchical routing is the emergence of controllable sparsity, which we leverage for efficient generation. As shown in Fig.~\ref{fig_teaser} (a) top and Sec.~\ref{sec:method_efficiency}, routing naturally identifies low-significance control pathways that can be skipped without degrading performance. We exploit this property through a budgeted training and inference scheme, combining action-aware distillation with significance-driven execution. This enables the model to selectively discard unnecessary computation, achieving substantial efficiency gains. Empirically, our efficient variant reduces latency by 47.1\% while maintaining strong control accuracy (Fig.~\ref{fig_teaser} (a) bottom).

To support this task, we further construct KASA, a new benchmark for \textbf{K}inematic \textbf{A}rticulated \textbf{S}urgical \textbf{A}ction. As illustrated in Fig.~\ref{fig_teaser} (b) and Sec.~\ref{sec:dataset_curation}, KASA provides curated annotations that bridge real surgical videos and articulated control signals. Our pipeline combines human-in-the-loop segmentation with differentiable pose tracking, enabling the extraction of part-aware geometry and temporally consistent trajectories without requiring external robot logs. The dataset covers diverse surgical actions such as knotting, needle grasping, and puncture, capturing both large-scale motion and fine-grained manipulation, and serves as a realistic testbed for evaluating structured control.

Extensive experiments (Sec.~\ref{sec:exp}) demonstrate that KVLR consistently improves action faithfulness, visual fidelity, and generalization. As summarized in Fig.~\ref{fig_teaser} (a) bottom, our method achieves 26.6\% Chamfer Distance decrease and 38.4\% FID decrease compared to strong baselines. KVLR consistently outperforms text-conditioned, raw-action-conditioned, and dense-visual-conditioned methods. Furthermore, KVLR shows favorable zero-shot transfer on unseen scenes.\looseness=-1 

In summary, this work makes the following contributions: (1) A kinematic-to-visual lifting paradigm that transforms articulated kinematics into unified, image-aligned control modalities; (2) A hierarchical routing method that dynamically allocates control across modalities and motion scales; (3) An action-adaptive efficient generation scheme that leverages routing-induced sparsity for significant computational acceleration; (4) KASA, a new benchmark with articulated annotations from real surgical videos for evaluating action-conditioned generation; (5) Comprehensive experiments demonstrating improved action faithfulness, visual fidelity, efficiency, and generalization.

\section{Related Work}
\subsection{Surgical Video Generation}
Surgical video generation has recently attracted growing attention for simulation, training, and predictive modeling in robotic surgery~\cite{min2025innovating,cho2024surgen,sakref2026empowering,ma2025open,xu2026generalized,turkcan2025towards,rapuri2026saw,chen2025far,ozawa2021synthetic,caballero2026generative,li2024llava,li2026surgpub,jeon2025surgen,wang2021towards,ross2023new,chen2018surgical,zhai2025generation,georgenthum2025enhancing,lee2019segmentation,jin2024surgical,koju2025surgical,eckhoff2023sages,wang2025accelerating,koksal2024sangria,wang2025surgvidlm,temsah2025openai,kato2025disturbance,loukas2018video,wang2025video,reiley2010motion,zhao2024see,venkatesh2025mitigating,maack2026approach,martyniak2025simuscope,yang2024automated,yin2026artificial,chen2025h,sun2024bora,li2024endosparse,cao2024medical,wu2025generation,li2024artificial,yang2025medical,fu2026colodiff,wang2025feat,sivakumar2025sg2vid,liu2025endogen,fu2026depthpilot,chen2025llama,yeganeh2025movis,wu2025learning,kurt2025modeling}. 
Early progress mainly focuses on scaling surgical video corpora and improving visual realism. 
MedGen~\cite{wang2025medgen} and Ophora~\cite{li2025ophora} build large-scale captioned datasets for generalist and ophthalmic video synthesis, while Endora~\cite{li2024endora} studies endoscopic video generation as a simulator for minimally invasive scenes. 
Simulation-assisted efforts such as SimuScope~\cite{Martyniak2024SimuScopeRE} combine surgical simulation with diffusion-based image translation to improve synthetic endoscopic realism.
Beyond visual synthesis, recent works explore controllable surgical video generation. 
SurGen~\cite{cho2024surgen} and Open-world~\cite{ma2025open} rely on language guidance for flexible scene generation, while SurgSora~\cite{chen2025surgsora} introduces structured RGBD-flow decomposition for controllable surgical video synthesis. 
HieraSurg~\cite{biagini2025hierasurg} further exploits hierarchy-aware segmentation to improve surgical video generation. 
More recent surgical world-modeling efforts move toward action-conditioned synthesis: SAW~\cite{rapuri2026saw} conditions video diffusion on lightweight signals such as language, reference scenes, tissue affordance masks, and 2D tool-tip trajectories, while SurgWorld~\cite{he2025surgworld} uses generative world modeling and pseudo-kinematics to support surgical policy learning. Open-H-Embodiment~\cite{Nelson2026OpenH} also highlights the importance of synchronized medical robotic videos and kinematics for embodied surgical AI.
Despite these advances, existing methods still primarily rely on coarse text prompts, 2D trajectory cues, or dense visual priors such as masks, depth, or optical flow. 
These controls are either insufficient for fine-grained articulated manipulation or costly to obtain at inference time. 
In contrast, our work targets \emph{action-conditioned} surgical video generation from sparse articulated controls. 
By lifting kinematics into a pixel-aligned action field and hierarchically routing conditioning across physical modalities and motion scales, KVLR aims to improve control precision, visual fidelity, and efficient inference under accessible robotic action inputs.\looseness=-1

\subsection{Controllable Generation with Structured Signals}
Beyond surgery, controllable video generation has been actively studied in robotics, autonomous driving, and general scene modeling~\cite{zheng2024opensora,yang2024cogvideox,gao2026pam,yang2025orv,jin2024lvsm,li2025rig3r,liang2025worldlens,yang2025resim,li2025uniscene,chen2025unimlvg,yanginstadrive,xu2025ad,ge2025unraveling,yang2025drivingview,guo2025dist,kong20253d,hu2025vision,liang2025worldlens}. In robotics, IRASim~\cite{zhu2024irasim} studies action-to-video prediction for robotic manipulation, while Cosmos-Transfer~\cite{alhaija2025cosmos} and RoboTransfer~\cite{liu2025robotransfer} condition synthesis on scene-level geometric cues such as depth or surface normals. In driving and embodied scene generation, occupancy-centric approaches use structured scene priors~\cite{li2025uniscene,li2025occscene,li2025scaling,li2025omninwm}, while MagicDrive~\cite{gao2023magicdrive} and Vista~\cite{gao2025vista} exploit structured multi-view or latent control for controllable synthesis. Related efforts in controllable rendering and 3D-aware generation also rely on dense scene representations or explicit geometric guidance~\cite{lin2025controllable,zhangjie2025difix3d+,liang2025diffusionrenderer}. These approaches demonstrate the value of structured conditioning, but they typically depend on dense maps, occupancy fields, or other heavy intermediate representations that are difficult to access in robotic surgery, where sensing is limited and occlusion is frequent~\cite{yang2025orv,zhangjie2025difix3d+,li2025uniscene,liang2025diffusionrenderer}. Differently, our method focuses on sparse articulated kinematic actions as the primary control interface, and in transforming them into pixel-aligned visual control through hierarchical kinematic-to-visual routing rather than relying on dense geometric priors.\looseness=-1

\section{Methodology}

\begin{figure}
    \centering
    \includegraphics[width=0.99\linewidth]{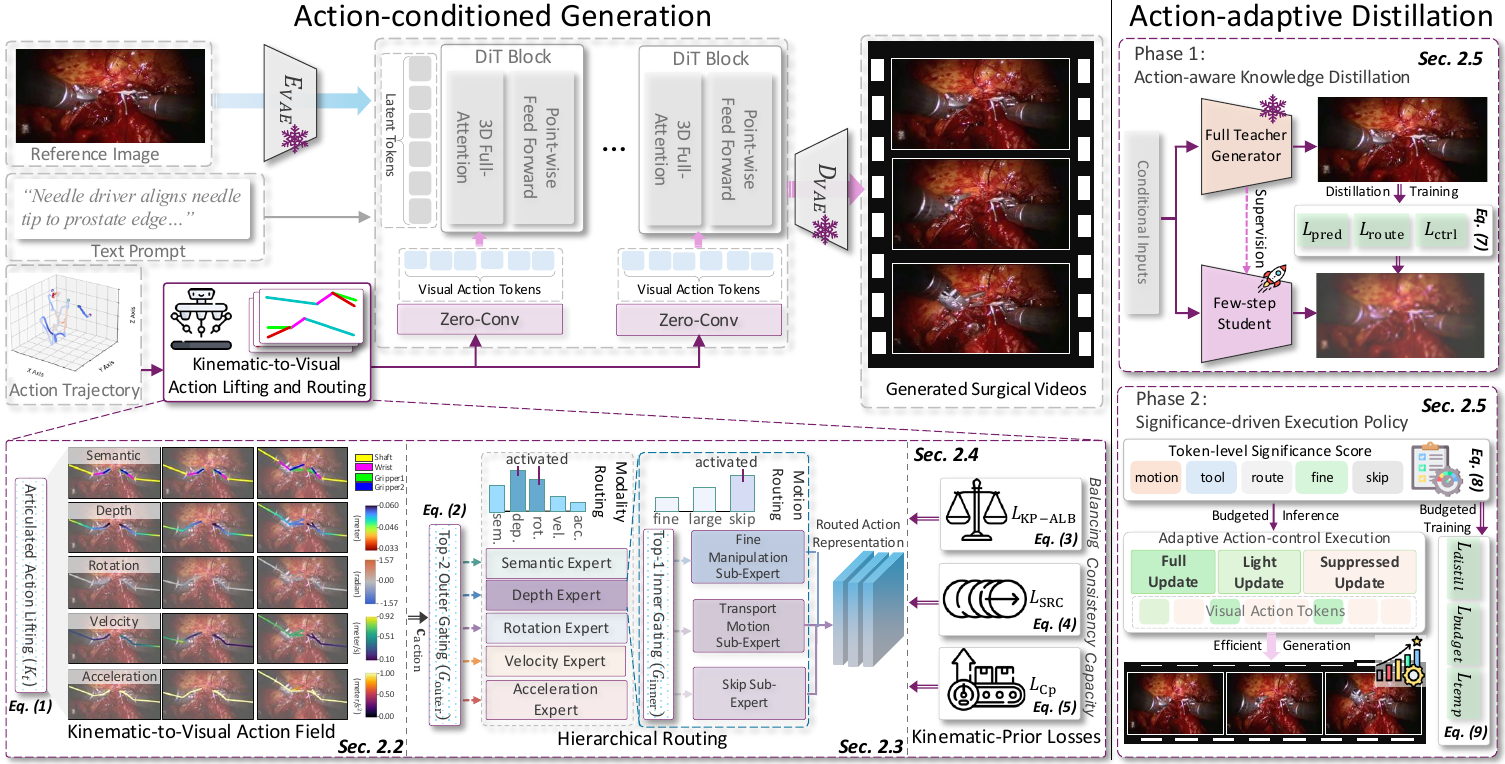}
    \caption{Architecture overview. Articulated kinematics are lifted into a pixel-aligned KVA-Field, then hierarchically routed across physical modalities and motion scales. This paradigm improves action faithfulness under heterogeneous dynamics and supports action-adaptive efficient generation.\looseness=-1}
    \label{fig_overall}
\end{figure}

\subsection{Framework Overview}
\label{sec:problem}

As shown in Fig.~\ref{fig_overall} top left, given a reference image $I^{\mathrm{ref}}$, a text prompt $y$, and an articulated action sequence $\mathrm{a}_{1:T}$ from the surgical robot, our goal is to generate a video $\hat{\mathrm{x}}_{1:T}$ conditioned on the specified surgical actions:
$
\hat{\mathrm{x}}_{1:T} = \mathcal{F}(I^{\mathrm{ref}}, y, \mathrm{a}_{1:T}).$ 
As illustrated in Fig.~\ref{fig_overall} bottom left, we first lift articulated kinematics into a pixel-aligned \emph{K}inematic-to-\emph{V}isual \emph{A}ction \emph{Field} (\emph{KVA-Field}), which encodes multi-modal cues directly in image space (Sec.~\ref{sec:kva_field}). 
We then apply \emph{hierarchical routing} to allocate conditional computation according to physical modalities and motion scales (Sec.~\ref{sec:routing}).
The kinematic-prior-guided routing loss functions are further designed to ensure physically meaningful, temporally stable, and efficient expert utilization (Sec.~\ref{sec:routing_learning}).
This design reveals intrinsic conditional sparsity, which is later leveraged for action-adaptive efficient generation (Sec.~\ref{sec:method_efficiency}). 

\subsection{Kinematic-to-Visual Action Field}
\label{sec:kva_field}

To bridge the gap between low-dimensional control and dense visual synthesis, we lift articulated instrument actions into a pixel-aligned \emph{Kinematic-to-Visual Action Field (KVA-Field)} with camera intrinsic and extrinsic parameters. We model the surgical tool with a part-aware articulated structure composed of a shaft, a wrist, and two gripper components (as shown in Fig.~\ref{fig_overall} bottom left). 
For each frame $t$, the articulated action is represented as
$[p_t, r_t, q^{\mathrm{sw}}_t, q^{\mathrm{lg}}_t, q^{\mathrm{rg}}_t] \in \mathbb{R}^{9}.
$
Here, $p_t \in \mathbb{R}^{3}$ and $r_t \in \mathbb{R}^{3}$ denote the wrist translation and wrist orientation, respectively, with $r_t$ expressed in axis-angle form. The scalar variables $q^{\mathrm{sw}}_t$, $q^{\mathrm{lg}}_t$, and $q^{\mathrm{rg}}_t$ denote the shaft-to-wrist, wrist-to-left-gripper, and wrist-to-right-gripper joint states. Together, these variables are turned into a compact articulated action representation $K_t\in\mathbb{R}^{H\times W\times 9},$ with channels corresponding to part semantics, depth, orientation, velocity, and acceleration:
\begin{equation}
K_t(h,w)=\left[s_t(h,w),d_t(h,w),\rho_t(h,w),v_t(h,w),\alpha_t(h,w)\right].
\end{equation}

Here, $s_t\in\mathbb{R}^{3}$ denotes part-aware semantic channels for shaft, wrist, and gripper; $d_t\in\mathbb{R}$ is rendered depth; $\rho_t\in\mathbb{R}$ is a local orientation descriptor; $v_t\in\mathbb{R}^{3}$ is projected velocity; and $\alpha_t\in\mathbb{R}$ is acceleration magnitude. $v_t$ and $\alpha_t$ are computed from consecutive articulated states: $
v_t =
\frac{\Phi(M_t)-\Phi(M_{t-1})}{\Delta t},$
$\alpha_t =
\frac{\|v_t-v_{t-1}\|_2}{\Delta t}.
\label{eq:motion_channels}
$ 
Here, $M_t$ is the articulated tool state and 
$\Phi(\cdot)$ denotes its camera-projected feature representation on the image plane.
\textbf{The resulting field encodes not only where the instrument is, but also how it is oriented and how it moves.} By explicitly separating semantic, geometric, and dynamic factors in image space, the KVA-Field provides the structured representation on top of which our hierarchical routing allocates conditional computation.

\subsection{Hierarchical Routing}
\label{sec:routing}

{To account for the spatial heterogeneity and heavy-tailed distribution of surgical motion, we introduce the \textit{Hierarchical Routing strategy}. Unlike standard Mixture-of-Experts (MoE)~\cite{masoudnia2014mixture,zhou2022mixture,rodriguez2026lar} architectures that route based solely on token content, our design employs a two-tier hierarchy.}
As shown in Fig.~\ref{fig_overall} bottom left, this architecture decouples the routing decision into two complementary axes: (1) \textit{Which physical modality is dominant at this spatial location?} (Tier 1), and (2) \textit{What scale of motion dynamics should be extracted within that modality?} (Tier 2). 
This hierarchical disentanglement allows the model to adaptively allocate computational resources to complex instrument dynamics.

\noindent\textbf{Tier 1: Multi-modality routing.} {The outer layer consists of $5$ modality-specific expert branches $\mathcal{E}_{\text{mod}} = \{E_i\}_{i=1}^5$. Each modality expert $E_i$ is structurally constrained to process only its corresponding physical channels from the 9-channel input KVA-Field ${K_t}$.
As shown in Fig.~\ref{fig_overall} bottom left, routing at this level is governed by an outer gating network $G_{\text{outer}}$. \todo{We encode ${K_t}$ with a lightweight shared action module
$\mathcal{A}_{\mathrm{act}}$ to obtain a holistic gating feature
$\mathbf{c}_{\text{action}}=\mathcal{A}_{\mathrm{act}}(K_t)$.
$G_{\text{outer}}$ is then computed as:}}\looseness=-1

\begin{equation}
    G_{\text{outer}}(\mathbf{c}_{\text{action}}, t) = \text{Softmax}\left( \text{Conv}_{1\times1}\left( \text{Concat}[\text{Pool}(\mathbf{c}_{\text{action}}), \tau(t)] \right) \right),
\end{equation}

where $\text{Pool}$ and $\tau(t)$ denote spatial average pooling and the diffusion timestep embedding, respectively.  
{To ensure training stability,} we employ a {scheduling} strategy~\cite{masoudnia2014mixture,zhou2022mixture} that transitions from dense softmax fusion to sparse top-$k$ routing (default $k=2$).  

\noindent\textbf{Tier 2: Motion-scale routing.}
{As illustrated in Fig.~\ref{fig_overall} bottom left, within each modality expert, we introduce a secondary layer to handle diverse motion dynamics. Each modality branch is composed of three sub-experts controlled by a top-1 inner gating mechanism $G_{\text{inner}}$:} \textbf{(1)} Fine Manipulation Sub-Expert: Utilizes $1\times1$ and dilated $3\times3$ convolutions to capture high-frequency details required for precision tasks like knotting or grasping. 
\textbf{(2)} Transport Motion Sub-Expert: Employs large-kernel ($7\times7$) depthwise convolutions and global context pooling to model fast tool sweeps and large displacements.
\textbf{(3)} Skip Sub-Expert: An identity mapping branch that allows zero-cost computation for static instrument regions, preventing unnecessary feature transformation.
 
\noindent\textbf{Routing fusion.} The inner router uses a $1\times1$ convolution to map the modality-specific feature vector at each spatial location to motion-scale logits. 
This produces an independent fine, large, or skip decision for each token, allowing the selected operator to adapt to the local kinematic state. 
The output of each modality branch is the selected sub-expert feature weighted by its inner-routing confidence. The final routed action representation is obtained by fusing the selected expert outputs.\looseness=-1

\begin{figure}[!t]
    \centering
    \includegraphics[width=0.99\linewidth]{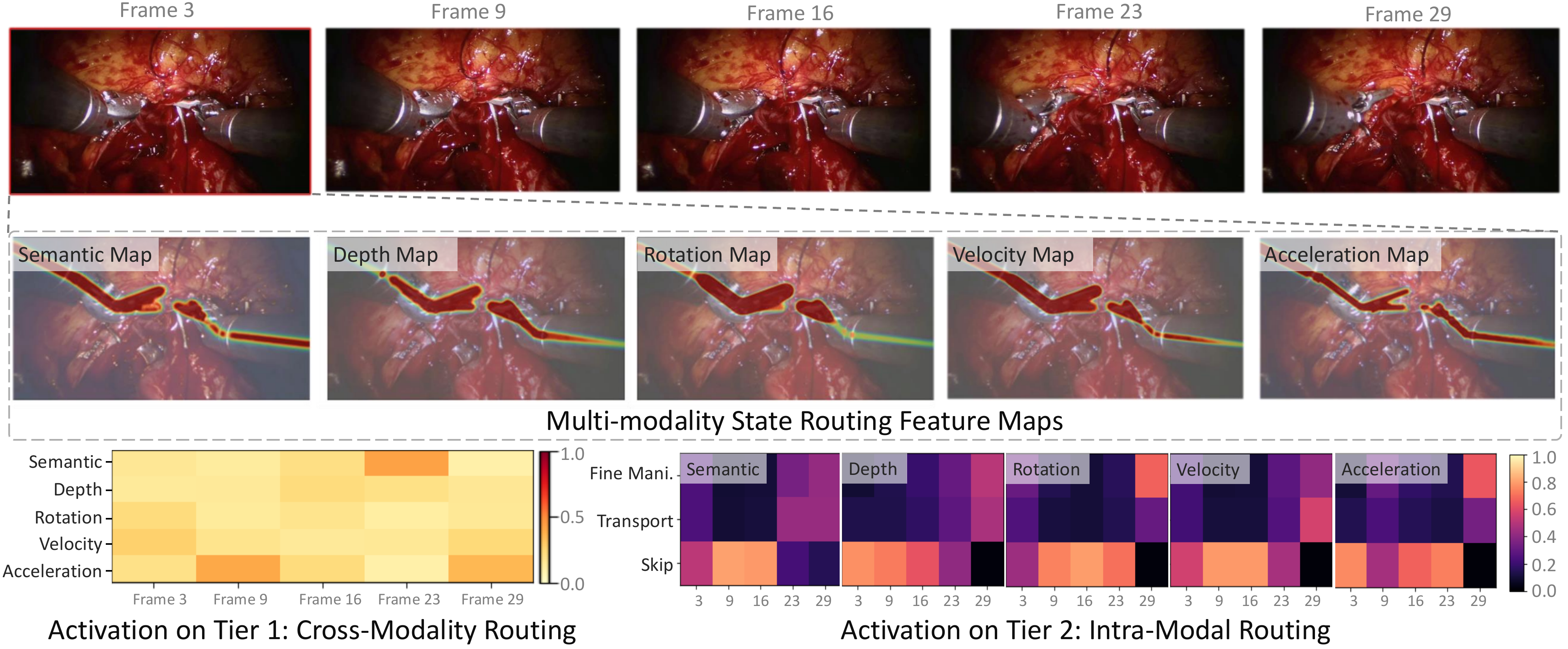}
    \vspace{-7pt}
    \caption{{Visualization of hierarchical routing.}
    We render the lifted action cues and show the learned Tier-1 modality and Tier-2 motion-scale activations. The routing patterns vary across motion phases, indicating action-dependent allocation of expert computation.
    }
    \label{fig:hierarchical_expert}
    \vspace{-17pt}
\end{figure}

\noindent\textbf{Hierarchical injection.}
 We then inject the fused control features into the video generator through zero-initialized additive layers on top of the DiT backbone, as shown in Fig.~\ref{fig_overall} top left. This preserves the pretrained generative prior at initialization while allowing the model to progressively learn action conditioning. In this way, the routing strategy allocates conditional computation according to both physical modality and motion complexity.
\todo{Fig.~\ref{fig:hierarchical_expert} shows that the learned routing is motion-dependent: low-motion regions favor semantic and skip pathways, while larger motions increasingly activate dynamic modalities and transport-oriented computation, matching the design of hierarchical routing.}\looseness=-1

\subsection{Kinematic-Prior Learning Loss Functions}
\label{sec:routing_learning}

The routing module in Sec.~\ref{sec:routing} should reflect the physical structure of surgical motion rather than purely statistical expert usage. However, as illustrated in Tab.~\ref{tab:ablation_loss}, previous MoE~\cite{ruby2020binary,fei2024scaling,zheng2025dense2moe} objectives, such as uniform load balancing, are poorly matched to the heavy-tailed and heterogeneous dynamics of articulated manipulation. We therefore introduce \emph{kinematic-prior learning} loss functions to make hierarchical routing physically meaningful, temporally stable, and compatible with efficient inference.\looseness=-1

\noindent\textbf{Kinematic-prior adaptive load balancing.}
Instead of enforcing uniform expert utilization, we guide the modality router with a dynamic physical signal derived from the KVA-Field. 
The \textbf{motivation} is that different frames are dominated by different physical factors: (1) static tool regions may be dominated by semantic and depth cues, (2) wrist articulation by rotation cues, and (3) rapid tool motion by velocity or acceleration cues. 
Thus, the target expert usage should depend on the physical signal present in the current action field rather than being fixed across all samples. 
Let $K_t^{\mathrm{sem}}\in\mathbb{R}^{H\times W\times 3}$, $K_t^{\mathrm{dep}}\in\mathbb{R}^{H\times W}$, $K_t^{\mathrm{rot}}\in\mathbb{R}^{H\times W}$, $K_t^{\mathrm{vel}}\in\mathbb{R}^{H\times W\times 3}$, and $K_t^{\mathrm{acc}}\in\mathbb{R}^{H\times W}$ denote the feature fields of depth, semantic, rotation, velocity and acceleration channels of the KVA-Field $K_t$ at frame $t$. 
For each spatial token $u\in\Omega$, where $\Omega$ is the image-token domain, we compute modality-wise physical magnitudes as
$
e_t(u)=
\left[
\|K_t^{\mathrm{sem}}(u)\|_2,\,
|K_t^{\mathrm{dep}}(u)|,\,
|K_t^{\mathrm{rot}}(u)|,\,
\|K_t^{\mathrm{vel}}(u)\|_2,\,
|K_t^{\mathrm{acc}}(u)|
\right]^\top .
$
These terms respectively measure part-presence strength, camera-depth magnitude, articulated rotation magnitude, projected motion magnitude, and acceleration magnitude. 
We aggregate them over spatial tokens to obtain the frame-level physical energy vector
$
\mathcal{E}_i(K_t)=\frac{1}{|\Omega|}\sum_{u\in\Omega} e_{t,i}(u), 
\ i\in\mathcal{M},
$
where $\mathcal{M}=\{\mathrm{sem},\mathrm{dep},\mathrm{rot},\mathrm{vel},\mathrm{acc}\}$ is the set of modality experts. 

The normalized physical signal is then
$
\pi_i(K_t)=
\frac{\mathcal{E}_i(K_t)}
{\sum_{j\in\mathcal{M}}\mathcal{E}_j(K_t)}.
$
This signal defines the desired routing mass for each modality expert according to the physical content of the current action field.
Let $P_i$ be the mean soft routing probability assigned to expert $i$, and let $f_i$ be the fraction of tokens whose top-ranked routing assignment selects expert $i$. 
Following sparse MOE load estimation, we define the realized routing load as $\ell_i=f_iP_i$ and minimize:\looseness=-1
\begin{equation}
\mathcal{L}_{\mathrm{KP\text{-}ALB}}
=
\sum_{i\in\mathcal{M}}
\left(
\ell_i-\pi_i(K_t)
\right)^2 .
\end{equation}

This objective encourages the router to allocate capacity according to physically meaningful action cues, rather than forcing all modality experts to be used uniformly.

\paragraph{Spatiotemporal routing consistency.}
Existing MoE diffusion models process tokens within a batch-level global pool, ignoring the temporal correlation inherent in sequential video frames~\cite{fei2024scaling,zheng2025dense2moe}. 
We introduce a Spatiotemporal Routing Consistency (SRC) loss to enforce temporal coherence in the continuous routing distributions, localized to moving instruments:
\begin{equation}
\mathcal{L}_{\mathrm{SRC}}=
\frac{1}{T-1}\sum_{t=1}^{T-1}\sum_{h,w}
M^{(t)}_{\mathrm{tool}}(h,w)\,
\left\|\mathrm{R}_t(h,w)-\mathrm{R}_{t-1}(h,w)\right\|_2^2,
\end{equation}

where $\mathrm{R} \in \mathbb{R}^{H \times W \times N}$ denotes the continuous routing probabilities at frame $t$, and ${M}_{\text{tool}}^{(t)} \in \{0, 1\}^{H \times W}$ is the binary spatial mask indicating surgical tool presence (derived from $K_t^{\mathrm{sem}}$). By masking the loss to instrument regions, we avoid penalizing routing changes in static background areas while stabilizing expert selection for dynamic tools.

\paragraph{Capacity predictor loss.}
To enable efficient inference without computing full top-$k$ routing at runtime, we train a lightweight {capacity predictor}~\cite{masoudnia2014mixture,zhou2022mixture} to anticipate routing decisions. The predictor receives detached action features and outputs raw logits $\hat{{R}}$ for each expert. We optimize via Binary Cross-Entropy against the actual routing mask ${R}^*$ generated by the gating networks:
\begin{equation}
\mathcal{L}_{\mathrm{CP}}
=
\mathrm{BCE}\left(\sigma(\hat{{R}}), \mathrm{sg}({R}^{*})\right),
\end{equation}
where $\sigma(\cdot)$ is the sigmoid function. During inference, the predictor's outputs are thresholded using EMA-updated per-expert quantiles, enabling dynamic routing with $\mathcal{O}(1)$ complexity per spatial token. The thresholds are updated every training step to track the evolving routing distribution, preventing expert starvation from positive-feedback loops. \todo{More implementation details are provided in Sec.~\ref{app:detailed_kinematic_prior}.}

\noindent\textbf{Overall objective.}
The final objective combines the generation loss with the above routing priors:
\begin{equation}
\mathcal{L}_{\mathrm{total}}
=
\mathcal{L}_{\mathrm{Flow}}
+\lambda_1 \mathcal{L}_{\mathrm{KP-ALB}}
+\lambda_2 \mathcal{L}_{\mathrm{SRC}}
+\lambda_3 \mathcal{L}_{\mathrm{CP}},
\end{equation}

where $\mathcal{L}_{\text{Flow}}$ is the standard flow-matching loss~\cite{lipman2022flow}, and $\lambda_{1:3}$ are weighting coefficients tuned as $\lambda_1=0.01$, $\lambda_2=0.005$, $\lambda_3=0.01$. \todo{More details on each loss function are presented in Sec.~\ref{app:detailed_kinematic_prior}.}\looseness=-1

\subsection{Action-adaptive Efficient Generation}
\label{sec:method_efficiency}

The structured control pathway improves action faithfulness and reveals where dense conditional computation is necessary. We exploit this property to build an \emph{action-adaptive efficient generation} scheme.\looseness=-1

\noindent\textbf{Phase 1: Action-aware distillation.}
As shown in Fig.~\ref{fig_overall} top right, we first distill a few-step student generator from the full teacher under the same conditioning inputs. To preserve action-aware behavior, the student is supervised not only on the final prediction, but also on the internal control structure:
\begin{equation}
\mathcal{L}_{\mathrm{distill}}
=
\mathcal{L}_{\mathrm{pred}}
+\lambda_r \mathcal{L}_{\mathrm{route}}
+\lambda_c \mathcal{L}_{\mathrm{ctrl}},
\end{equation}

where $\mathcal{L}_{\mathrm{pred}}$ matches the teacher prediction, $\mathcal{L}_{\mathrm{route}}$ aligns routing behavior, and $\mathcal{L}_{\mathrm{ctrl}}$ preserves structured action-control features. 
{More details on each loss function are provided in Sec.~\ref{sec:app_efficiency}.
This transfers the teacher's control policy into a few-step generator instead of distilling appearance alone.\looseness=-1}

\noindent\textbf{Phase 2: Significance-driven execution.}
We then derive a token-level action significance score from the lifted action field and routing signals:
\begin{equation}
s(u)
=
w_m e_{{\mathrm{motion}}}(u)
+
w_t m_{{\mathrm{tool}}}(u)
+
w_r c_{{\mathrm{route}}}(u)
+
w_f q_{{\mathrm{fine}}}(u)
-
w_s p_{{\mathrm{skip}}}(u),
\label{eq:significance}
\end{equation}

where $u$ denotes a latent token; 
$e_{{\mathrm{motion}}}$, $m_{{\mathrm{tool}}}$, $c_{{\mathrm{route}}}$, $q_{{\mathrm{fine}}}$, and $p_{{\mathrm{skip}}}$ denote normalized motion intensity, tool presence, routing confidence, fine-motion preference, and skip probability.
The non-negative weights $w_m,w_t,w_r,w_f,w_s$ control the relative contribution of each term.
At inference, $s(u)$ partitions latent tokens into three execution modes. As shown in Fig.~\ref{fig_overall} bottom-right, high-significance tokens perform a full control update by evaluating the KVA encoder, routers, and selected experts.
Medium-significance tokens perform a light update by reusing cached routing decisions and only recomputing the final residual adapter.
Low-significance tokens suppress the current control update and reuse the previous cached control residual.
A frame-level refresh interval (detailed in Sec.~\ref{sec:appendix_refresh}) prevents cache drift by forcing full updates at fixed temporal intervals.\looseness=-1

\noindent\textbf{Budget-aware objective.}
We optimize the efficient path with:
\begin{equation}
\mathcal{L}_{\mathrm{eff}}
=
\mathcal{L}_{\mathrm{distill}}
+\lambda_b \mathcal{L}_{\mathrm{budget}}
+\lambda_t \mathcal{L}_{\mathrm{temp}},
\end{equation}

where $\mathcal{L}_{\mathrm{budget}}$ enforces a target compute budget and $\mathcal{L}_{\mathrm{temp}}$ stabilizes reused or lightly updated regions over time. This objective reduces generation cost while preserving action-conditioned control. {More details are provided in the Sec.~\ref{sec:app_efficiency} of the appendix.\looseness=-1}

\begin{figure*}[!t]
\centering
\includegraphics[width=0.99\linewidth]{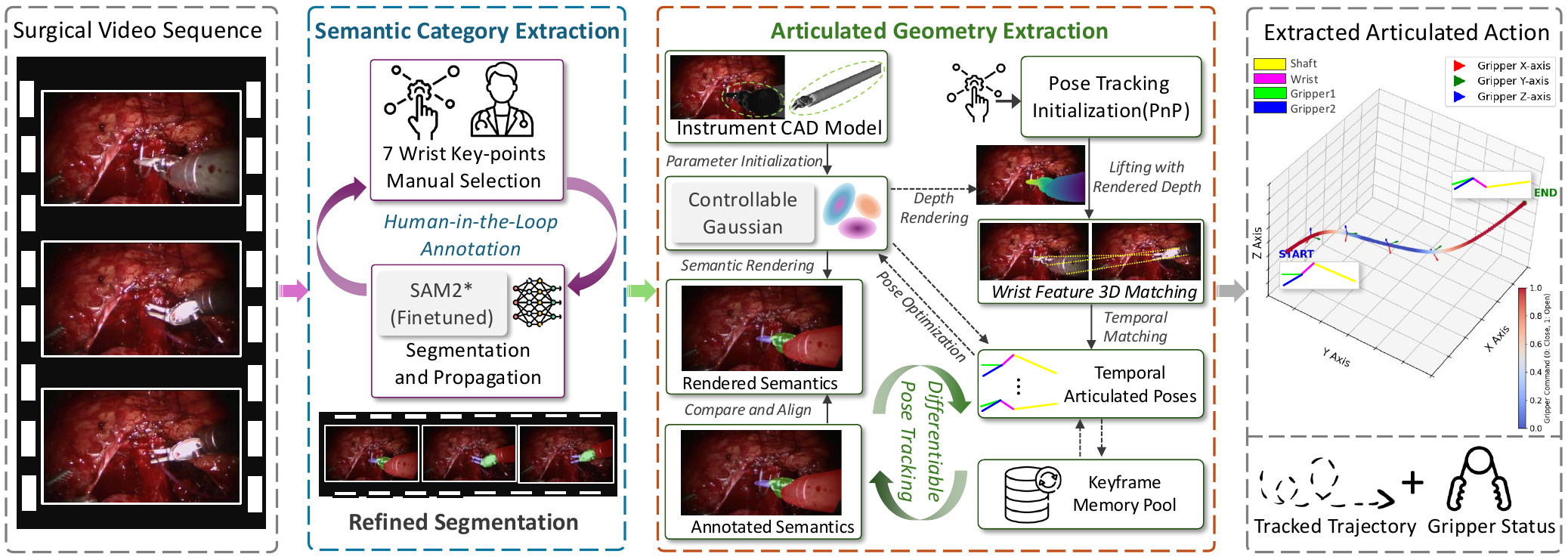}
\vspace{-9pt}
\caption{Data construction pipeline. Given robotic surgical videos, we obtain articulated action supervision through human-in-the-loop semantic annotation and differentiable pose tracking, producing lightweight yet visually grounded control signals for action-conditioned video generation.}
\label{fig_dataset}
\vspace{-16pt}
\end{figure*}

\section{KASA Benchmark Construction}
\label{sec:dataset_curation}
\noindent\textbf{Benchmark overview.}
A key obstacle in action-conditioned surgical video generation is the lack of benchmarks pairing real surgical videos with explicit articulated controls. Existing works~\cite{cho2024surgen,he2026cosmosh,li2025ophora,chen2025surgsora,biagini2025hierasurg,rapuri2026saw} typically provide coarse language descriptions or dense visual annotations, but not lightweight action supervision aligned with visual generation. We construct the Kinematic Action-centric Surgical (\textit{KASA}) benchmark for articulated tool-action-conditioned generation, associating each video with video-reconstructed annotations. These compact, visually grounded controls are compatible with robotic settings and aligned with structured control signals for kinematic-to-visual action modeling.\looseness=-1

\noindent\textbf{Task-driven sequence selection.}
We curate KASA from the SAR-RARP dataset~\cite{psychogyios2023sar}, collecting 1,047 surgical scenes with 105,175 frames. The benchmark focuses on three representative sub-actions of \emph{needle grasping}, \emph{needle puncture}, and \emph{knotting}, which together cover a broad range of surgical dynamics, from large transport motion to fine local manipulation. Videos are downsampled to 30 FPS, and each sequence contains 55-235 frames. 
KASA exhibits articulated motion, frequent self-occlusion, and complex instrument-tissue interactions, providing a challenging test platform for evaluating both action faithfulness and structured control under heterogeneous surgical dynamics.

\noindent\textbf{Human-in-the-loop semantic annotation.}
To obtain reliable part-aware supervision, we adopt a Human-in-the-loop Semantic Annotation (HSA) procedure (Fig.~\ref{fig_dataset} left). We first estimate part-level masks for the instrument shaft, wrist, and grippers using a SAM2 model~\cite{ravi2024sam} finetuned on SAR-RARP, and then refine the propagated predictions through interactive human correction. This step removes part confusion, boundary leakage, and temporal inconsistency, yielding clean semantic labels that are crucial for disentangling different physical factors in our structured control representation.\looseness=-1

\noindent\textbf{Differentiable pose tracking.}
To recover continuous articulated geometry, we further introduce Differentiable Pose Tracking (DPT), which estimates temporally consistent pose trajectories directly from videos without relying on external kinematic logs or hardware tracking (Fig.~\ref{fig_dataset} right). DPT uses a render-and-compare formulation initialized from instrument CAD priors and aligns rendered part-aware geometry with the observed video evidence over time. The resulting articulated trajectories provide the geometric and motion supervision, enabling downstream construction of semantic, depth, rotation, and motion-aligned control signals for action-conditioned video generation.
To further validate KASA, Sec.~\ref{sec_independence} and Sec.~\ref{app:kas_construction} report independent-evaluator checks and annotation quality validations.\looseness=-1

\section{Experiment}\label{sec:exp}

\subsection{Experimental Setup}
\noindent\textbf{Training.} We instantiate our models based on OpenSora-2.0~\cite{zheng2024opensora}. The models are trained on NVIDIA H20 GPUs with a batch size of 24 using AdamW~\cite{loshchilov2017decoupled}. The learning rate is set to $1\times10^{-5}$ with weight decay of 0.01. 
We split KASA into 85\% training and 15\% validation sets. 
The videos are generated at a default resolution of $576\times1024$ with clip lengths aligned with the ground-truth sequences.\looseness=-1 

\noindent\textbf{Baselines.} 
Our full model uses 50 denoising steps, while \textit{KVLR-fast} denotes the distilled efficient variant with a 4-step student. We compare with Wan-2.2~\cite{wan2025}, HunyuanVideo~\cite{kong2024hunyuanvideo}, OpenSora-2.0~\cite{zheng2024opensora}, Cosmos-H-Surgical~\cite{he2026cosmosh}, and SurgSora~\cite{chen2025surgsora}, all finetuned on KASA under the same split and input protocol. Models marked with `$^*$' use ControlNet-style injection (same parameter scale as ours) with raw-action (articulated trajectory vectors) conditions, while models with `$^\dagger$' leverage dense visual conditions as SurgSora~\cite{chen2025surgsora} (\textit{i.e.}, depth, semantic, and flow maps).
The results are reported with the mean and standard deviation over three runs.
Additional details are provided Sec.~\ref{sec:app_impl} of the appendix.\looseness=-1


\subsection{Main Results}\label{main_results}

\begin{table}[!t]
\centering
\caption{{Action faithfulness on KASA.}
All methods are finetuned under the same split.\looseness=-1 
}
\vspace{-5pt}
\label{tab_faithful}
    \renewcommand{\arraystretch}{1.01} 
    \setlength{\tabcolsep}{7.1pt} 
    \begin{threeparttable}
    \resizebox{\textwidth}{!}{%
        \begin{tabular}{l *{12}{c}} 
            \toprule[1pt]
            \multirow{2}{*}{Method}
            & \multicolumn{3}{c}{\cellcolor{green!5}Knotting} 
            & \multicolumn{3}{c}{\cellcolor{blue!5}NeedleGrasping} 
            & \multicolumn{3}{c}{\cellcolor{brown!5}NeedlePuncture} \\
            \cmidrule(lr){2-4} \cmidrule(lr){5-7} \cmidrule(lr){8-10}
            & CD $\downarrow$ & TI $\uparrow$  & AF $\downarrow$ 
            & CD $\downarrow$ & TI $\uparrow$  & AF $\downarrow$ 
            & CD $\downarrow$ & TI $\uparrow$  & AF $\downarrow$  \\
            \midrule[1pt]
            \rowcolor{gray!10}
            \multicolumn{10}{c}{\textit{Text-conditioned Generation}} \\
            Wan-2.2     & $169.35${\tiny$\pm3.12$} & $0.56${\tiny$\pm0.02$} & $6.54${\tiny$\pm0.45$}  & $135.80${\tiny$\pm2.85$} & $0.57${\tiny$\pm0.02$} & $26.95${\tiny$\pm1.20$} & $98.15${\tiny$\pm2.15$}  & $0.65${\tiny$\pm0.02$} & $5.31${\tiny$\pm0.38$}  \\
            HunyuanVideo & $162.81${\tiny$\pm2.95$} & $0.57${\tiny$\pm0.01$} & $6.07${\tiny$\pm0.50$}  & $131.63${\tiny$\pm2.60$} & $0.59${\tiny$\pm0.02$} & $26.05${\tiny$\pm1.15$} & $94.11${\tiny$\pm1.95$}  & $0.67${\tiny$\pm0.01$} & $4.82${\tiny$\pm0.35$}  \\
            OpenSora-2.0 & $173.42${\tiny$\pm3.40$} & $0.53${\tiny$\pm0.02$} & $7.02${\tiny$\pm0.62$}  & $140.56${\tiny$\pm3.10$} & $0.55${\tiny$\pm0.02$} & $28.35${\tiny$\pm1.40$} & $102.67${\tiny$\pm2.45$} & $0.63${\tiny$\pm0.02$} & $5.72${\tiny$\pm0.45$}  \\
            Cosmos-H-Surgical & $150.26${\tiny$\pm2.80$} & $0.59${\tiny$\pm0.02$} & $4.88${\tiny$\pm0.35$}  & $119.38${\tiny$\pm2.45$} & $0.61${\tiny$\pm0.02$} & $23.64${\tiny$\pm1.05$} & $85.30${\tiny$\pm1.80$}  & $0.71${\tiny$\pm0.01$} & $3.44${\tiny$\pm0.25$}  \\
            \midrule[0.5pt]
            \rowcolor{gray!10}
            \multicolumn{10}{c}{\textit{Raw-action-conditioned Generation}} \\
            Wan-2.2$^*$   & $135.48${\tiny$\pm2.55$} & $0.68${\tiny$\pm0.02$} & $5.23${\tiny$\pm0.40$}  & $108.64${\tiny$\pm2.15$} & $0.70${\tiny$\pm0.02$} & $21.56${\tiny$\pm0.95$} & $78.52${\tiny$\pm1.65$}  & $0.78${\tiny$\pm0.02$} & $4.25${\tiny$\pm0.30$} \\
            HunyuanVideo$^*$ & $130.25${\tiny$\pm2.40$} & $0.70${\tiny$\pm0.01$} & $4.86${\tiny$\pm0.35$}  & $105.31${\tiny$\pm2.05$} & $0.72${\tiny$\pm0.01$} & $20.84${\tiny$\pm0.85$} & $75.29${\tiny$\pm1.50$}  & $0.80${\tiny$\pm0.01$} & $3.86${\tiny$\pm0.25$}  \\
            OpenSora-2.0$^*$ & $138.74${\tiny$\pm2.65$} & $0.65${\tiny$\pm0.02$} & $5.62${\tiny$\pm0.45$}  & $112.45${\tiny$\pm2.25$} & $0.68${\tiny$\pm0.02$} & $22.68${\tiny$\pm1.10$} & $82.14${\tiny$\pm1.75$}  & $0.76${\tiny$\pm0.02$} & $4.58${\tiny$\pm0.35$} \\
            Cosmos-H-Surgical$^*$ 
            & $120.21${\tiny$\pm2.34$} & $0.74${\tiny$\pm0.01$} & $3.91${\tiny$\pm0.22$}
            & $95.51${\tiny$\pm1.86$} & $0.76${\tiny$\pm0.02$} & $18.91${\tiny$\pm0.74$}
            & $68.24${\tiny$\pm1.41$} & $0.84${\tiny$\pm0.01$} & $2.75${\tiny$\pm0.18$} \\ 
                        \midrule[1pt]
            \rowcolor{gray!10}
            \multicolumn{10}{c}{\textit{Visual-action-conditioned Generation}} \\
            
            SurgSora$^\dagger$ 
            & $119.50${\tiny$\pm2.10$} & $0.75${\tiny$\pm0.02$} & $3.85${\tiny$\pm0.20$} 
            & $94.80${\tiny$\pm1.65$} & $0.77${\tiny$\pm0.01$} & $18.50${\tiny$\pm0.65$} 
            & $67.50${\tiny$\pm1.50$} & $0.85${\tiny$\pm0.02$} & $2.65${\tiny$\pm0.15$} \\
            OpenSora-2.0$^\dagger$ 
            & $113.80${\tiny$\pm1.95$} & $0.84${\tiny$\pm0.02$} & $1.51${\tiny$\pm0.15$} 
            & $91.10${\tiny$\pm1.80$} & $0.87${\tiny$\pm0.02$} & $12.20${\tiny$\pm0.55$} 
            & $61.40${\tiny$\pm1.35$} & $0.87${\tiny$\pm0.01$} & $1.88${\tiny$\pm0.12$} \\
            KVLR-fast(ours)
            & $114.32${\tiny$\pm1.91$} & $0.85${\tiny$\pm0.01$} & $1.54${\tiny$\pm0.14$}
            & $91.85${\tiny$\pm1.48$} & $0.88${\tiny$\pm0.01$} & $12.45${\tiny$\pm0.51$}
            & $63.45${\tiny$\pm1.12$} & $0.86${\tiny$\pm0.01$} & $1.95${\tiny$\pm0.12$} \\
            KVLR(ours)
            & $\mathbf{110.57}${\tiny$\mathbf{\pm1.62}$} & $\mathbf{0.88}${\tiny$\mathbf{\pm0.01}$} & $\mathbf{1.18}${\tiny$\mathbf{\pm0.10}$}
            & $\mathbf{89.22}${\tiny$\mathbf{\pm1.35}$} & $\mathbf{0.91}${\tiny$\mathbf{\pm0.01}$} & $\mathbf{10.67}${\tiny$\mathbf{\pm0.43}$}
            & $\mathbf{60.90}${\tiny$\mathbf{\pm0.96}$} & $\mathbf{0.89}${\tiny$\mathbf{\pm0.01}$} & $\mathbf{1.59}${\tiny$\mathbf{\pm0.09}$} \\
            \bottomrule[1pt]
        \end{tabular}%
    }
    \end{threeparttable}
    \vspace{-8pt}
\end{table}

\begin{table}[!t]
\centering
\caption{{Generation Fidelity on KASA.} 
PSNR/SSIM are averaged over generated frames, FID is computed on generated frames as images, and FVD on 17-frame clips.}
\vspace{-5pt}
\label{tab_video_gen}
\renewcommand{\arraystretch}{1.0}
\setlength{\tabcolsep}{1.0pt}
\begin{threeparttable}
\resizebox{1.03\textwidth}{!}{
\begin{tabular}{l *{12}{c}} 
            \toprule[1pt]
            \multirow{2}{*}{Method}
            & \multicolumn{4}{c}{\cellcolor{green!5}Knotting} 
            & \multicolumn{4}{c}{\cellcolor{blue!5}NeedleGrasping} 
            & \multicolumn{4}{c}{\cellcolor{brown!5}NeedlePuncture} \\
            \cmidrule(lr){2-5} \cmidrule(lr){6-9} \cmidrule(lr){10-13}
            & PSNR $\uparrow$ & SSIM $\uparrow$  & FID $\downarrow$ & FVD $\downarrow$  
            & PSNR $\uparrow$ & SSIM $\uparrow$  & FID $\downarrow$ & FVD $\downarrow$  
            & PSNR $\uparrow$ & SSIM $\uparrow$  & FID $\downarrow$ & FVD $\downarrow$ \\
            \midrule[1pt]
            \rowcolor{gray!10}
            \multicolumn{13}{c}{\textit{Text-conditioned Generation}} \\
            Wan-2.2    & $16.20${\tiny$\pm0.25$} & $0.55${\tiny$\pm0.02$} & $45.30${\tiny$\pm2.40$} & $420.50${\tiny$\pm18.50$}  & $15.30${\tiny$\pm0.22$} & $0.52${\tiny$\pm0.02$} & $48.20${\tiny$\pm2.60$} & $450.30${\tiny$\pm20.10$} & $16.80${\tiny$\pm0.26$} & $0.58${\tiny$\pm0.02$} & $40.10${\tiny$\pm2.10$} & $310.40${\tiny$\pm15.50$}  \\
            HunyuanVideo & $16.50${\tiny$\pm0.22$} & $0.57${\tiny$\pm0.01$} & $42.10${\tiny$\pm2.15$} & $395.20${\tiny$\pm16.20$}  & $15.60${\tiny$\pm0.20$} & $0.54${\tiny$\pm0.01$} & $44.50${\tiny$\pm2.35$} & $415.80${\tiny$\pm18.40$} & $17.10${\tiny$\pm0.24$} & $0.60${\tiny$\pm0.01$} & $37.20${\tiny$\pm1.95$} & $285.20${\tiny$\pm14.20$}  \\
            OpenSora-2.0 & $16.10${\tiny$\pm0.28$} & $0.54${\tiny$\pm0.02$} & $47.50${\tiny$\pm2.55$} & $445.80${\tiny$\pm19.80$}  & $15.10${\tiny$\pm0.25$} & $0.51${\tiny$\pm0.02$} & $50.40${\tiny$\pm2.80$} & $470.20${\tiny$\pm22.50$} & $16.60${\tiny$\pm0.25$} & $0.56${\tiny$\pm0.02$} & $42.30${\tiny$\pm2.25$} & $335.60${\tiny$\pm16.80$}  \\
            Cosmos-H-Surgical & $17.20${\tiny$\pm0.20$} & $0.62${\tiny$\pm0.01$} & $36.50${\tiny$\pm1.90$} & $325.40${\tiny$\pm14.50$}  & $16.40${\tiny$\pm0.18$} & $0.59${\tiny$\pm0.01$} & $38.50${\tiny$\pm2.10$} & $355.50${\tiny$\pm15.80$} & $17.80${\tiny$\pm0.22$} & $0.65${\tiny$\pm0.01$} & $31.50${\tiny$\pm1.75$} & $245.70${\tiny$\pm12.40$}  \\
            \midrule[0.5pt]
            \rowcolor{gray!10}
            \multicolumn{13}{c}{\textit{Raw-action-conditioned Generation}} \\
            Wan-2.2$^*$  & $17.80${\tiny$\pm0.24$} & $0.66${\tiny$\pm0.02$} & $31.60${\tiny$\pm1.85$} & $280.20${\tiny$\pm13.50$}  & $16.90${\tiny$\pm0.21$} & $0.63${\tiny$\pm0.02$} & $33.10${\tiny$\pm2.05$} & $310.80${\tiny$\pm14.80$} & $18.20${\tiny$\pm0.25$} & $0.68${\tiny$\pm0.02$} & $26.30${\tiny$\pm1.65$} & $200.50${\tiny$\pm11.50$} \\
            HunyuanVideo$^*$ & $18.10${\tiny$\pm0.22$} & $0.68${\tiny$\pm0.01$} & $28.40${\tiny$\pm1.75$} & $255.40${\tiny$\pm12.80$}  & $17.30${\tiny$\pm0.19$} & $0.65${\tiny$\pm0.01$} & $30.20${\tiny$\pm1.95$} & $285.90${\tiny$\pm13.60$} & $18.60${\tiny$\pm0.23$} & $0.71${\tiny$\pm0.01$} & $23.40${\tiny$\pm1.55$} & $175.40${\tiny$\pm10.80$}  \\
            OpenSora-2.0$^*$ & $17.50${\tiny$\pm0.26$} & $0.64${\tiny$\pm0.02$} & $34.20${\tiny$\pm2.10$} & $305.60${\tiny$\pm15.20$}  & $16.60${\tiny$\pm0.24$} & $0.61${\tiny$\pm0.02$} & $36.40${\tiny$\pm2.25$} & $335.70${\tiny$\pm16.50$} & $17.90${\tiny$\pm0.27$} & $0.66${\tiny$\pm0.02$} & $29.10${\tiny$\pm1.85$} & $225.80${\tiny$\pm12.20$} \\
            Cosmos-H-Surgical$^*$
            & $18.90${\tiny$\pm0.23$} & $0.72${\tiny$\pm0.01$} & $24.46${\tiny$\pm1.81$} & $225.72${\tiny$\pm12.35$} 
            & $18.20${\tiny$\pm0.18$} & $0.70${\tiny$\pm0.01$} & $25.87${\tiny$\pm1.94$} & $251.26${\tiny$\pm15.18$}
            & $19.14${\tiny$\pm0.20$} & $0.75${\tiny$\pm0.01$} & $19.72${\tiny$\pm1.67$} & $142.97${\tiny$\pm10.74$} \\ 
            \midrule[1pt]
            \rowcolor{gray!10}
            \multicolumn{13}{c}{\textit{Visual-action-conditioned Generation}} \\
            
            SurgSora$^\dagger$ 
            & $18.85${\tiny$\pm0.21$} & $0.73${\tiny$\pm0.02$} & $24.15${\tiny$\pm1.50$} & $222.40${\tiny$\pm11.50$}
            & $18.35${\tiny$\pm0.15$} & $0.71${\tiny$\pm0.01$} & $25.40${\tiny$\pm1.65$} & $248.50${\tiny$\pm13.20$}
            & $19.25${\tiny$\pm0.18$} & $0.76${\tiny$\pm0.01$} & $19.35${\tiny$\pm1.45$} & $140.60${\tiny$\pm9.80$} \\
            OpenSora-2.0$^\dagger$ 
            & $20.75${\tiny$\pm0.25$} & $0.79${\tiny$\pm0.01$} & $16.80${\tiny$\pm1.60$} & $168.90${\tiny$\pm10.80$}
            & $19.90${\tiny$\pm0.20$} & $0.75${\tiny$\pm0.02$} & $18.95${\tiny$\pm1.75$} & $185.20${\tiny$\pm14.50$}
            & $20.75${\tiny$\pm0.15$} & $0.83${\tiny$\pm0.01$} & $12.90${\tiny$\pm1.30$} & $97.50${\tiny$\pm8.90$} \\
            KVLR-fast(ours)
            & $20.45${\tiny$\pm0.17$} & $0.78${\tiny$\pm0.01$} & $18.25${\tiny$\pm1.12$} & $185.32${\tiny$\pm10.46$}
            & $19.68${\tiny$\pm0.15$} & $0.76${\tiny$\pm0.01$} & $20.15${\tiny$\pm1.18$} & $205.40${\tiny$\pm12.33$}
            & $20.55${\tiny$\pm0.16$} & $0.81${\tiny$\pm0.01$} & $14.60${\tiny$\pm0.82$} & $112.50${\tiny$\pm7.64$} \\
            KVLR(ours)
            & $\mathbf{21.09}${\tiny$\mathbf{\pm0.14}$} & $\mathbf{0.82}${\tiny$\mathbf{\pm0.01}$} & $\mathbf{14.78}${\tiny$\mathbf{\pm0.89}$} & $\mathbf{154.56}${\tiny$\mathbf{\pm8.72}$}
            & $\mathbf{20.17}${\tiny$\mathbf{\pm0.13}$} & $\mathbf{0.80}${\tiny$\mathbf{\pm0.01}$} & $\mathbf{16.60}${\tiny$\mathbf{\pm0.91}$} & $\mathbf{168.94}${\tiny$\mathbf{\pm9.85}$}
            & $\mathbf{21.22}${\tiny$\mathbf{\pm0.12}$} & $\mathbf{0.85}${\tiny$\mathbf{\pm0.01}$} & $\mathbf{11.75}${\tiny$\mathbf{\pm0.68}$} & $\mathbf{87.10}${\tiny$\mathbf{\pm5.96}$} \\
            \bottomrule[1pt]
        \end{tabular}%
 
}
\end{threeparttable}
\vspace{-16pt}
\end{table}

\noindent\textbf{Action faithfulness.}
We evaluate Chamfer Distance (CD), Temporal IoU (TI), and Area Flicker (AF) using an independently trained Grounded SAM~\cite{ren2024grounded}, comparing predicted tool masks with ground-truth action masks; details are in Sec.~\ref{sec:appendix_metrics}.
As shown in Tab.~\ref{tab_faithful}, text-only baselines struggle to follow surgical actions, and raw-action injection improves control but remains limited.
Dense-visual-condition baselines further improve action precision using semantic, depth, and flow maps, but such inputs are often unavailable in robotic settings and still underperform KVLR.
\textit{KVLR-fast} retains compelling control fidelity, showing that action alignment is largely preserved under reduced computation.\looseness=-1

\begin{figure}[!t]
    \centering
    \includegraphics[width=0.99\linewidth]{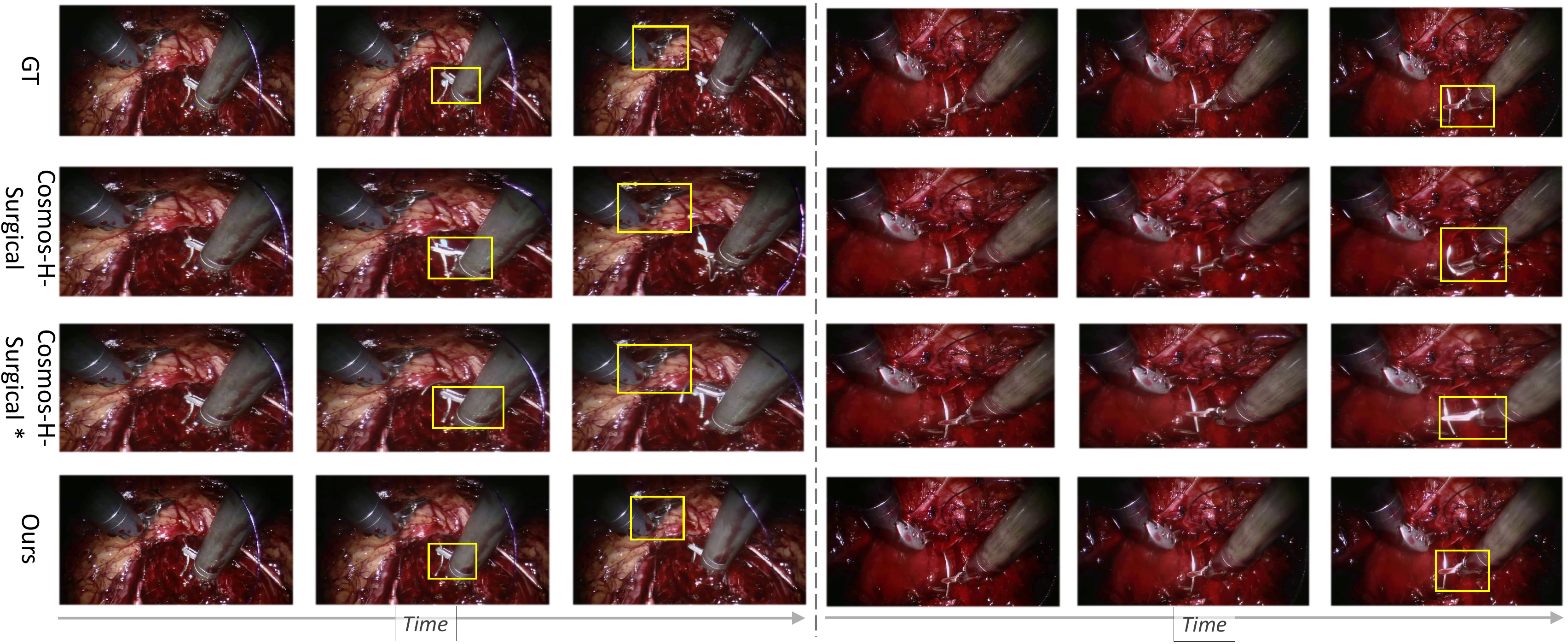}
    \caption{Qualitative comparison on KASA. Under cluttered scenes and poor illumination, KVLR better preserves tool states and action-consistent evolution than Cosmos-H-Surgical$^*$.\looseness=-1}
    \label{fig:qual_compare}
\end{figure}

\begin{figure}[!t]
    \centering
    \includegraphics[width=0.99\linewidth]{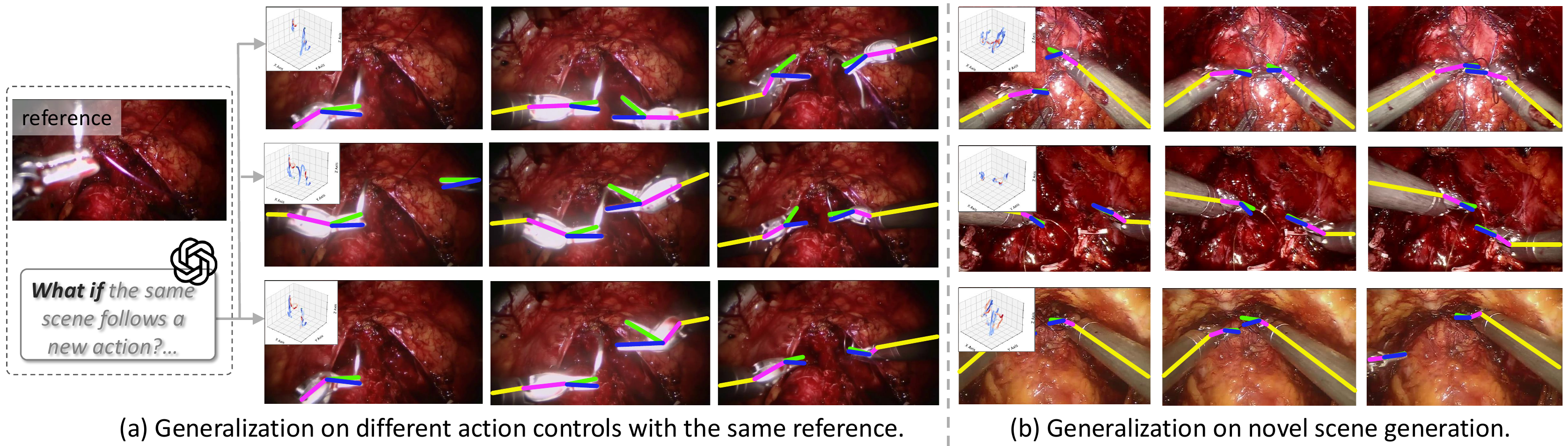}
    \caption{{Generalization results.}
    (a) Controllable synthesis under user-specified actions with the same reference.
    (b) Qualitative transfer to unseen scenes with different appearances and motion patterns.\looseness=-1}
    \label{fig:generalization}
\end{figure}

\noindent\textbf{Generation fidelity.}
Tab.~\ref{tab_video_gen} shows consistent gains in visual fidelity.
Direct action injection improves over text-only generation, while structured visual-action-condition baselines further benefit from dense scene-level guidance.
KVLR achieves the best overall performance, particularly on \textit{Knotting} and \textit{NeedleGrasping}, where realistic tool-tissue evolution requires accurate local motion control.
This suggests that kinematic-to-visual lifting and routing provide benefits beyond simply introducing dense visual conditions.
Fig.~\ref{fig:qual_compare} further qualitatively confirms that KVLR better preserves action-consistent evolution and interaction outcomes under cluttered scenes and poor illumination.\looseness=-1

\begin{wraptable}{r}{0.55\linewidth} 
\centering
\vspace{-20pt}
\caption{{Computational efficiency evaluation.} }
\vspace{-0pt}
\renewcommand\tabcolsep{8.0pt}
\resizebox{\linewidth}{!}{
\begin{tabular}{l|cc|cc}
\toprule
\multirow{2}{*}{Method} & \multicolumn{2}{c|}{288$\times$512} & \multicolumn{2}{c}{576$\times$1024} \\
\cmidrule(lr){2-3} \cmidrule(lr){4-5}
 & \makecell{Latency (s)$\downarrow$} & \makecell{FLOPs$\downarrow$ \\ (PF/clip)} & \makecell{Latency (s)$\downarrow$}  & \makecell{FLOPs$\downarrow$ \\ (PF/clip)} \\
\midrule
 Wan-2.2$^*$          & 2.38 & 3.42  & 11.85 & 12.18  \\
 HunyuanVideo$^*$     & 6.12 & 9.15  & 30.54 & 32.45  \\
 Cosmos-H-Surgical$^*$           & 0.70 & 1.91  & 4.32  & 7.55   \\  
 SurgSora$^\dagger$   & 3.85 & 5.42  & 18.25 & 19.65  \\
 OpenSora-2.0$^\dagger$& 5.12 & 7.55  & 25.60 & 26.85  \\
\midrule
 KVLR (Ours)          & 4.64 & 6.94  & 23.39 & 24.77  \\  
 KVLR-fast (Ours)     & \textbf{0.37} & \textbf{0.55} & \textbf{1.88} & \textbf{1.98} \\ 
\bottomrule
\end{tabular}}
\label{tab_efficiency}
\end{wraptable}

\noindent\textbf{Computational efficiency.}
Tab.~\ref{tab_efficiency} reports latency and FLOPs on H20 GPUs with 17-frame clips. Full KVLR prioritizes action faithfulness and fidelity, whereas KVLR-fast targets efficient inference. At both resolutions,
KVLR-fast achieves the lowest latency and FLOPs while retaining strong action alignment, showing
that routing-derived significance exposes useful sparsity for adaptive acceleration.\looseness=-1

\begin{table}[!t]
\centering
\caption{Cross-dataset generalization on GraSP.}
\label{tab:grasp_generalization}
\small
\setlength{\tabcolsep}{10.0pt}
\resizebox{0.8\linewidth}{!}{
\begin{tabular}{lcccccc}
\toprule
Method & CD$\downarrow$ & TI$\uparrow$ & AF$\downarrow$ & PSNR$\uparrow$ & FID$\downarrow$  \\
\midrule
Cosmos-H-Surgical$^*$ & 134.8\tiny$\pm$3.9 & 0.68\tiny$\pm$0.03 & 8.42\tiny$\pm$0.58 & 16.9\tiny$\pm$0.3 & 28.3\tiny$\pm$1.8   \\
SurgSora$^\dagger$ & 127.5\tiny$\pm$3.4 & 0.71\tiny$\pm$0.02 & 7.15\tiny$\pm$0.45 & 17.6\tiny$\pm$0.2 & 24.0\tiny$\pm$1.5  \\
OpenSora-2.0$^\dagger$ & 115.3\tiny$\pm$3.1 & 0.77\tiny$\pm$0.01 & 4.78\tiny$\pm$0.38 & 18.9\tiny$\pm$0.2 & 17.0\tiny$\pm$1.3  \\
KVLR-fast (Ours) & 117.6\tiny$\pm$2.8 & 0.76\tiny$\pm$0.02 & 5.05\tiny$\pm$0.31 & 18.7\tiny$\pm$0.2 & 19.1\tiny$\pm$1.1   \\
{KVLR (Ours)} & \textbf{111.9\tiny$\pm$2.5} & \textbf{0.79\tiny$\pm$0.01} & \textbf{4.28\tiny$\pm$0.26} & \textbf{19.2\tiny$\pm$0.1} & \textbf{16.7\tiny$\pm$0.9}   \\
\bottomrule
\end{tabular}}
\end{table}

\noindent\textbf{Cross-dataset generalization.}
We evaluate zero-shot transfer on GraSP~\cite{valderrama2020tapir,ayobi2023matis}, which differs in tissue appearance, illumination, and viewpoint.
Kinematics are extracted by our DPT pipeline without human correction, introducing realistic tracking noise.
Without fine-tuning, KVLR maintains favorable faithfulness and fidelity over the evaluated baselines (Tab.~\ref{tab:grasp_generalization}), suggesting improved robustness to domain shift.
This trend is consistent with the qualitative transfer results in Fig.~\ref{fig:generalization} (b) and the ``what-if'' action control examples in Fig.~\ref{fig:generalization} (a).\looseness=-1

\begin{wraptable}{r}{0.48\linewidth}
\centering
\caption{{Trajectory-level verification.} Ctr. and End. are centerline and distal-endpoint errors in pixels; Dir. is motion-direction accuracy(\%).}
\label{tab:evaluator_independence}
\renewcommand{\arraystretch}{1.0}
\setlength{\tabcolsep}{15pt}
\resizebox{\linewidth}{!}{
\begin{tabular}{lccc}
\toprule
Method & Ctr.$\downarrow$ & End.$\downarrow$ & Dir.$\uparrow$ \\
\midrule
Cosmos-H-Surgical$^*$ & 11.83 & 15.62 & 71.4 \\
KVLR-fast (Ours) & 8.05 & 10.24 & 82.1 \\
KVLR (Ours) & \textbf{7.21} & \textbf{9.13} & \textbf{84.8} \\
\bottomrule
\end{tabular}}
\end{wraptable}
\noindent\textbf{Trajectory-level verification.}\label{sec_independence}
Rather than only computing the mask-overlap metrics, we provide an additional trajectory-level evaluation that extracts the visible tool centerline and distal endpoint from each generated video using an independently trained Grounded SAM-based tracker~\cite{ren2024grounded}, and compares them with the projected trajectory induced by the reference articulated action. 
As shown in Tab.~\ref{tab:evaluator_independence}, KVLR reduces both centerline and endpoint errors and improves motion-direction accuracy over Cosmos-H-Surgical$^*$, indicating that its gains are not specific to the default mask-overlap protocol. 
Action-perturbation checks in appendix~\ref{app:evaluation_independence} further confirm that the generated motion follows the specified action trajectory.\looseness=-1

\subsection{Ablation Study}

\noindent\textbf{Architecture design.}
Tab.~\ref{tab:ablation_core} ablates KVLR under parameter-matched control branches.
Despite the matched capacity, direct conditioning with raw action yields limited gains, suggesting that the bottleneck is the fundamental representational mismatch between low-dimensional articulated kinematics and dense visual evolution. 
Replacing raw actions with the KVA-Field bridges this gap, driving a substantial performance leap (\textit{e.g.}, CD drops to 96.45). 
Adding hierarchical routing dynamically allocates this fixed parameter budget across varying physical modalities and motion scales, further boosting action faithfulness without inflating latency, showing that action lifting and structured routing are complementary. 
\todo{Qualitative comparisons on different settings are provided in Fig.~\ref{supple_abl_arc}} of the appendix.\looseness=-1


\begin{table}[!t]
\centering
\caption{Architecture ablation under parameter-matched control branches.\looseness=-1}
\label{tab:ablation_core}
\renewcommand{\arraystretch}{1.12}
\setlength{\tabcolsep}{14pt}
\begin{threeparttable}
\resizebox{\textwidth}{!}{%
\begin{tabular}{lcccccc}
\toprule[1pt]
Method & Total Params & CD $\downarrow$ & TI $\uparrow$ & FID $\downarrow$ & PSNR $\uparrow$ & Latency (s) $\downarrow$ \\
\midrule[1pt]
Text-only baseline 
& 11.00B & 138.88 & 0.57 & 93.31 & 12.80 & \textbf{4.59} \\
+ Raw action vector + direct condition 
& 12.06B & 111.11 & 0.69 & 79.56 & 15.08 & 4.61 \\
+ KVA-Field + direct condition 
& 12.06B & 96.45  & 0.78 & 45.20 & 17.65 & 4.68 \\
+ KVA-Field + Tier-1 routing 
& 12.05B & 91.30  & 0.84 & 26.85 & 19.34 & 4.66 \\
+ KVA-Field + Tier-1 + Tier-2 routing 
& 12.06B 
& \textbf{86.89} & \textbf{0.89} & \textbf{14.37} & \textbf{20.82} & 4.64 \\
\bottomrule[1pt]
\end{tabular}%
}
\end{threeparttable}
\end{table}

\begin{wraptable}{r}{0.34\textwidth} 
    \centering
    \caption{{Kinematic-prior ablation.}}
    \label{tab:ablation_loss}
    \renewcommand{\arraystretch}{1.12}
    \setlength{\tabcolsep}{3.0pt}  
    \resizebox{0.99\linewidth}{!}{ 
        \begin{tabular}{lcccc}
        \toprule[1pt]
        Method & CD $\downarrow$ & TI $\uparrow$ & AF $\downarrow$ & FID $\downarrow$ \\
        \midrule[1pt]
        Full w/o KP-ALB & 94.30 & 0.85 & 5.20 & 21.65 \\
        Full w/o SRC    & 90.12 & 0.81 & 8.75 & 17.50 \\
        Full w/o CP     & 89.45 & 0.87 & 5.12 & 16.20 \\
        Full model      & \textbf{86.89} & \textbf{0.89} & \textbf{4.48} & \textbf{14.37} \\
        \bottomrule[1pt]
        \end{tabular} 
    }
\end{wraptable}

\noindent\textbf{Kinematic-prior losses.}
Tab.~\ref{tab:ablation_loss} shows that the three priors are complementary.
Replacing KP-ALB with uniform load balancing~\cite{ruby2020binary,fei2024scaling,zheng2025dense2moe} mainly hurts spatial alignment and fidelity, while removing SRC causes the largest drop in temporal stability; removing CP also degrades all metrics, indicating that predictable routing is important for stable control.
Overall, the best results require kinematic-prior-guided regularization throughout the routing process. \todo{More qualitative comparisons on the effect of different losses are provided in Fig.~\ref{supple_abl_loss} of the appendix.}\looseness=-1

\begin{wraptable}{r}{0.5\textwidth} 
\caption{{Efficient generation ablation}.}
\label{tab:ablation_efficiency}
\renewcommand{\arraystretch}{1.12}
\setlength{\tabcolsep}{1.2pt}
\begin{threeparttable}
\resizebox{\linewidth}{!}{%
\begin{tabular}{lcccc}
\toprule
Method & CD $\downarrow$     & FID $\downarrow$ & Latency $\downarrow$ & FLOPs $\downarrow$ \\
\midrule[1pt]
Teacher (full generator, 50 steps) 
& \textbf{86.89}  & \textbf{14.37} & 4.64 & 6.94 \\
\midrule[0.5pt]
Generic distillation (DMD2)  
& 98.45  & 19.82 & 0.52 & 0.78 \\
Action-aware student (few-step only) 
& 87.50  & 15.60 & 0.52 & 0.78  \\
\quad + spatial adaptive execution 
& 88.65   & 16.55 & 0.42 & 0.63\\
\quad + temporal cache (Final efficient) 
& 89.87   & 17.67 & \textbf{0.37} & \textbf{0.55}  \\
\bottomrule
\end{tabular}%
}
\end{threeparttable}
\end{wraptable}

\noindent\textbf{Efficient generation.}
Tab.~\ref{tab:ablation_efficiency} separates generic few-step acceleration from action-aware efficiency.
Under the same 4-step budget, DMD2~\cite{yin2024improved} reduces latency to 0.52\,s but degrades control, increasing CD from 86.89 to 98.45.
By contrast, our action-aware student achieves the same latency and FLOPs while preserving much stronger control fidelity, with CD 87.50.
Spatial adaptive execution further reduces latency/FLOPs to 0.42\,s/0.63\,PF by skipping low-significance control updates, and temporal caching reaches 0.37\,s/0.55\,PF with moderate degradation.
These results indicate that few-step distillation provides the main acceleration, while action-aware distillation and routing-derived execution better preserve control under reduced computation.
\todo{A more detailed breakdown analysis is provided in Sec.~\ref{sec:app_eff_breakdown}}.\looseness=-1

\vspace{-12pt}
\section{Conclusion} 
This work presents KVLR, a structured framework for action-conditioned surgical video generation from articulated inputs. 
KVLR lifts kinematics into a pixel-aligned KVA-Field and routes control across modalities and motion scales, improving measured action alignment and generation quality under complex surgical dynamics. 
We also introduce KASA, a benchmark with articulated annotations from real robotic surgical videos. 
Experiments show that KVLR outperforms evaluated baselines, while KVLR-fast reduces inference cost with largely preserved action control. 

\noindent\textbf{Limitations.} Generated videos should be viewed as research aids for simulation and data augmentation rather than clinical evidence. 
Additional discussions on limitations and broader impact are provided in Sec.~\ref{sec:appendix_lim} of the appendix.\looseness=-1

\bibliographystyle{plainnat}
\bibliography{main.bib}

\newpage
\appendix


\section*{\huge \centering Appendix}

\startcontents
{
    \hypersetup{linkcolor=black}
    \printcontents{}{1}{}
}

\newpage

\section{Formal Motivation}
\label{app:formal_motivation}

This section of the appendix provides a more discussions on the main design choices in KVLR.

\subsection{Why Pixel-Aligned Lifting Helps}
\label{app:lifting_motivation}

The first design choice in KVLR is to lift sparse articulated actions into a dense image-aligned control field before injecting them into the video generator. This subsection formalizes the intuition that raw-action conditioning must learn both spatial alignment and control-feature prediction, while the lifted field externalizes part of the alignment problem.

Let $u$ denote a spatial-temporal token and let $z_t^\star(u)$ be an ideal local control feature sufficient for action-conditioned generation at token $u$:
\begin{equation}
z_t^\star(u)=\phi^\star(\xi_t(u)),
\end{equation}
where $\xi_t(u)$ denotes the local physical state relevant to token $u$, such as part identity, depth, local orientation, projected motion, and motion change.

Under raw-action conditioning, the model predicts a token-wise control feature directly from the global articulated action:
\begin{equation}
\hat{z}^{\mathrm{raw}}_t(u)=g_{\mathrm{raw}}(a_t,u).
\end{equation}
This requires the learned branch to implicitly approximate both the projection from the global articulated state to token $u$ and the mapping from the resulting local physical state to generator features.\looseness=-1

In contrast, KVLR first constructs a pixel-aligned action field,
\begin{equation}
K_t(u)=L_u(a_t),
\end{equation}
where $L_u(\cdot)$ denotes the deterministic lifting operation at token $u$. The local control feature is then predicted as:\looseness=-1
\begin{equation}
\hat{z}^{\mathrm{lift}}_t(u)=g_{\mathrm{lift}}(K_t(u)).
\end{equation}
The lifted representation separates the spatial-alignment step from the learned control-feature mapping.

To make this intuition explicit, suppose that the lifted field recovers the relevant local physical state up to an abstract error $\varepsilon_{\mathrm{lift}}$, while a learned projection from raw actions recovers it up to $\varepsilon_{\mathrm{proj}}$. Also suppose that $\phi^\star$ is $L_{\phi}$-Lipschitz and that the learned mappings have approximation errors $\varepsilon^{\mathrm{lift}}_{\mathrm{app}}$ and $\varepsilon^{\mathrm{raw}}_{\mathrm{app}}$. A direct triangle-inequality decomposition gives:
\begin{equation}
\left\|
\hat{z}^{\mathrm{lift}}_t(u)-z_t^\star(u)
\right\|
\lesssim
L_{\phi}\varepsilon_{\mathrm{lift}}
+
\varepsilon^{\mathrm{lift}}_{\mathrm{app}},
\end{equation}
whereas raw-action conditioning yields:
\begin{equation}
\left\|
\hat{z}^{\mathrm{raw}}_t(u)-z_t^\star(u)
\right\|
\lesssim
L_{\phi}\varepsilon_{\mathrm{proj}}
+
\varepsilon^{\mathrm{raw}}_{\mathrm{app}}.
\end{equation}

These quantities are the decomposition that clarifies the design assumption behind KVLR: pixel-aligned lifting is expected to help when the hand-constructed action field reduces the spatial-alignment burden more than it restricts the representation. This is consistent with the parameter-matched ablation in the main paper, where replacing raw action vectors with the KVA-Field provides a substantial gain before adding hierarchical routing.

\subsection{When Hierarchical Routing Is Useful}
\label{app:routing_motivation}

The second design choice in KVLR is to route lifted action features across physical modalities and motion scales. We interpret this routing mechanism as an inductive bias rather than as a universally stronger function class. A sufficiently large monolithic conditioner can, in principle, approximate heterogeneous control mappings. The motivation for routing is that, under a fixed parameter and computation budget, surgical scenes contain spatial-temporal regimes with different dominant factors: static or low-motion regions often require semantic and geometric consistency, while high-motion regions require dynamic cues and larger motion operators.

Let $\kappa_t(u)$ denote an unobserved control regime for token $u$, and let $f^\star_{\kappa_t(u)}$ be the corresponding local control map. A monolithic conditioner approximates all regimes using a shared operator:
\begin{equation}
\hat{z}^{\mathrm{mono}}_t(u)=f_{\mathrm{mono}}(K_t(u)).
\end{equation}
A routed conditioner instead selects an expert or sub-expert according to the local control regime:
\begin{equation}
\hat{z}^{\mathrm{route}}_t(u)=f_{\hat{\kappa}_t(u)}(K_t(u)),
\end{equation}
where $\hat{\kappa}_t(u)$ denotes the selected routing branch.

If the average within-regime approximation error is denoted by $\bar{\varepsilon}$, and if the probability of selecting an unsuitable branch is denoted by $\eta$, then the routed error can be informally decomposed as
\begin{equation}
\mathbb{E}_{t,u}
\left[
\left\|
\hat{z}^{\mathrm{route}}_t(u)-z_t^\star(u)
\right\|
\right]
\lesssim
\bar{\varepsilon}+C\eta,
\end{equation}
where $C$ is a bounded mismatch constant measuring the cost of choosing an unsuitable expert. This expression separates two intuitive error sources: expert approximation error and routing error. It is not meant to provide an estimable guarantee, because $\bar{\varepsilon}$, $\eta$, and $C$ are not measured in the actual model.\looseness=-1

This decomposition motivates two aspects of KVLR. First, hierarchical routing can be useful when different physical regimes are easier to approximate with specialized operators than with a single shared pathway. Second, the benefit of specialization depends on routing quality, which motivates the kinematic-prior objectives introduced in the main method. In practice, the value of this inductive bias is evaluated empirically through the parameter-matched ablation comparing raw-action injection, KVA-Field injection, Tier-1 routing, and Tier-1 plus Tier-2 routing.

\subsection{Interpretation of Kinematic-Prior Regularization}
\label{app:prior_interpretation}

The kinematic-prior objectives are introduced to bias the router toward physically plausible and temporally stable assignments. They constrain observable aspects of routing behavior that are relevant to surgical motion.

\paragraph{Kinematic-prior adaptive load balancing.}
The KP-ALB objective aligns modality usage with the physical signal strength in the KVA-Field:
\begin{equation}
\mathcal{L}_{\mathrm{KP\mbox{-}ALB}}
=
\sum_{i \in \mathcal{M}}
\left(
\ell_i-\pi_i(K_t)
\right)^2,
\end{equation}
where $\pi_i(K_t)$ denotes the normalized physical prior of modality $i$, and $\ell_i$ denotes its routing load. Unlike uniform load balancing, this term does not force all experts to be used equally. Instead, it discourages routing patterns that ignore the relative strength of semantic, geometric, and dynamic action cues.

\paragraph{Spatiotemporal routing consistency.}
The SRC objective penalizes frame-to-frame routing variation around tool regions:
\begin{equation}
\mathcal{L}_{\mathrm{SRC}}
=
\frac{1}{T-1}
\sum_{t=1}^{T-1}
\sum_{h,w}
M^{(t)}_{\mathrm{tool}}(h,w)
\left\|
R_t(h,w)-R_{t-1}(h,w)
\right\|_2^2 .
\end{equation}
This term reflects the assumption that routing decisions around articulated tools should change smoothly unless the underlying motion changes. It reduces routing flicker but does not enforce temporal smoothness in static background regions.

\paragraph{Capacity prediction.}
The capacity-prediction objective trains a lightweight predictor to approximate routing decisions from detached action features:
\begin{equation}
\mathcal{L}_{\mathrm{CP}}
=
\mathrm{BCE}
\left(
\sigma(\hat{R}), \mathrm{sg}(R^\star)
\right).
\end{equation}
This makes routing decisions more predictable from action-derived signals, which later supports the efficient execution policy. Its role is practical rather than theoretical: it helps expose a stable approximation of where full control computation is likely to matter.

Overall, these regularizers reduce the degrees of freedom of the router and make the learned routing easier to interpret.

\section{Discussion on Limitations and Broader Impact}\label{sec:appendix_lim}

\subsection{Limitations}
Our framework is developed for action-conditioned surgical video generation under articulated robotic control, and is evaluated on representative suturing-related sub-actions with curated supervision. While the proposed structured kinematic-to-visual control improves faithfulness, fidelity, and efficiency in this setting, extending the framework to broader surgical procedures and more diverse instruments may require additional adaptation in data curation and model scaling. In addition, as with other generative models, quantitative improvements on benchmark metrics do not fully capture all aspects of clinical realism or downstream utility, and human-centered evaluation remains important for future study.

\subsection{Broader Impact}
This work has the potential to benefit surgical simulation, training, and data augmentation by enabling more controllable video generation from lightweight robotic action signals. At the same time, synthetic surgical videos should not be interpreted as a substitute for real clinical evidence or expert judgment. We do not advocate direct deployment of generated videos for autonomous clinical decision-making. More broadly, responsible use of such models requires appropriate attention to data governance, benchmarking transparency, and the intended educational or research context.\looseness=-1

\subsection{Asset License and Usage Note}
This work builds on several existing datasets, models, and software components, including SAR-RARP for source video data, SAM2 for segmentation-assisted annotation, OpenSora and related pretrained video-generation components for model initialization, and publicly available baseline methods used for comparison. We credit the original creators of these assets in the main paper and references, and use them in accordance with their respective licenses, terms of use, and research-only restrictions where applicable. For all external assets, we follow the usage conditions specified by their official releases or project pages. Any future release of our code, models, or derived benchmark assets will also be conditioned on compatibility with the licenses and usage terms of these underlying resources. We refer to the official releases/pages of these assets for the exact license terms and usage conditions, and will include explicit license identifiers in any public release package.

\section{More Implementation Details}
\label{sec:app_impl}

\subsection{Detailed Experimental Setup}
\label{app:exp_setup}

\noindent\textbf{Training details.}~\label{sec:appendix_training}
Unless otherwise specified, all models are trained on NVIDIA H20 GPUs with a batch size of 24 using AdamW~\cite{loshchilov2017decoupled}, with a learning rate of $1\times10^{-5}$ and weight decay of 0.01. 
We split the KASA benchmark into 85\% training data and 15\% validation data.
Our model is trained with a progressive two-stage protocol for stable convergence.
In the first stage, we finetune the generator using text-only conditioning to preserve the pretrained video prior and adapt the backbone to the surgical domain.
In the second stage, we continue training with joint text-and-action conditioning, enabling the model to learn structured action control on top of the adapted generator.
The overall training is implemented on 24 GPUs for approximately 7 days.
Unless otherwise stated, generated videos use a spatial resolution of $576\times1024$, and the generated sequence length is aligned with the corresponding ground-truth clip.
For inference, the full model uses 50 denoising steps.
The efficient variant, denoted as \textit{Ours-fast}, uses the distilled 4-step student described in Sec.~\ref{sec:method_efficiency}.\looseness=-1

\noindent\textbf{Baseline setup.}
We compare with Wan-2.2~\cite{wan2025}, HunyuanVideo~\cite{kong2024hunyuanvideo}, OpenSora-2.0~\cite{zheng2024opensora}, and Cosmos-H-Surgical~\cite{he2026cosmosh}.
All evaluated baselines are finetuned on KASA under the same train and validation split and the same input protocol for fair comparison.
For text-conditioned baselines, we use the original conditioning interface of each model.
For action-conditioned comparison, models marked with `$^*$' are augmented with ControlNet-style direct action injection, so that articulated actions are provided as explicit additional controls.
Models with `$^\dagger$' leverage dense visual conditions as SurgSora~\cite{chen2025surgsora}, including depth, semantic, and flow maps.
During inference, each baseline uses its default sampling setting when applicable.
Our method uses the default settings described above unless otherwise specified in the corresponding experiment.

\noindent\textbf{Evaluation metrics.}~\label{sec:appendix_metrics}
We evaluate three aspects: \emph{action faithfulness}, \emph{generation fidelity}, and \emph{computational efficiency}. 
For action faithfulness, each generated video is matched to its corresponding KASA annotation by video ID. 
Let $\Omega \subset \mathbb{Z}^{2}$ denote the image pixel domain. 
For an instrument instance $m$ at frame $t$, we render the articulated annotation into a ground-truth action tube mask $T_t^{m}\subset\Omega$ by projecting the recovered 3D instrument skeleton into the image plane and drawing a fixed-width tube around the projected skeleton. 
We also render a thin centerline mask $C_t^{m}$ and projected articulated keypoints for auxiliary validation. 
To extract the generated instrument region, SAM2 or Grounded SAM is initialized with the first valid ground-truth action tube mask and propagated through the generated video, producing a predicted binary mask $P_t^{m}\subset\Omega$ for each instrument instance and frame.

The reported Chamfer Distance (CD) measures the symmetric pixel distance between the generated instrument mask and the target action tube:
\begin{equation}
\mathrm{CD}_t^{m}
=
\frac{1}{2}
\left(
\frac{1}{|P_t^{m}|}\sum_{p\in P_t^{m}} d(p,T_t^{m})
+
\frac{1}{|T_t^{m}|}\sum_{g\in T_t^{m}} d(g,P_t^{m})
\right),
\end{equation}
where $|\cdot|$ denotes the number of foreground pixels and
$d(x,A)=\min_{a\in A}\|x-a\|_2$ is the Euclidean distance from pixel $x$ to the nearest foreground pixel in mask $A$, computed using distance transforms. Lower CD indicates that the generated instrument region is spatially closer to the projected target action.\looseness=-1

Temporal IoU (TI) evaluates frame-to-frame consistency of the generated tool region. 
Let $\bar{P}_t=\bigcup_m P_t^{m}$ be the union of predicted masks over all instrument instances at frame $t$. 
We compute:
\begin{equation}
\mathrm{TI}_t
=
\frac{|\bar{P}_t\cap \bar{P}_{t-1}|}
{|\bar{P}_t\cup \bar{P}_{t-1}|}.
\end{equation}
Higher TI indicates more temporally stable generated tool masks. 
Area Flicker (AF) measures relative frame-to-frame area variation:
\begin{equation}
\mathrm{AF}_t
=
\frac{\big||\bar{P}_t|-|\bar{P}_{t-1}|\big|}
{\max(|\bar{P}_{t-1}|,1)}.
\end{equation}
Lower AF indicates less abrupt temporal fluctuation in the generated action region. 
All action-faithfulness metrics are averaged over valid frames and instrument instances with non-empty projected action targets.

For generation fidelity, we report PSNR, SSIM, FID, and FVD.
PSNR and SSIM are averaged over all generated frames.
FID is computed by treating generated video frames as images.
FVD is computed on 17-frame clips to evaluate temporal realism and video-level consistency.

For efficiency, we report inference latency and FLOPs, and additionally report peak memory in the efficient-generation ablations.
Unless otherwise stated, all validation metrics are averaged over sequences.
The computational efficiency results in the main paper are measured on NVIDIA H20 GPUs using 17-frame clips.

Main benchmark tables report averages over the three action subsets, while uncertainty estimates are provided for the representative quantitative analyses in Sec.~\ref{app_add}.

\subsection{Backbone Instantiation and Training Context}
\label{sec:app_backbone}

Our generator $G$ is instantiated on top of OpenSora, which consists of a pretrained text encoder, a 3D VAE, and a trainable DiT backbone. In the main paper, we keep Sec.~\ref{sec:problem} intentionally compact and mainly describe the two-stage control decomposition. Here, we provide the omitted implementation context.\looseness=-1

Given a reference image $I^{\mathrm{ref}}$, a text prompt $y$, and an articulated action
sequence $\mathrm{a}_{1:T}$, we first encode the visual input into the latent space of the
pretrained 3D VAE. The text condition is processed by the pretrained text encoder, while the articulated action sequence is transformed into the Kinematic-to-Visual Action Field described in Sec.~\ref{sec:app_kva}. The resulting structured action features are then processed by the hierarchical routing module before being injected into the DiT backbone through a lightweight conditional branch.

Unless otherwise specified, we preserve the original latent dimensionality and temporal tokenization of the OpenSora backbone. The action branch is designed to be additive
rather than replacing the original visual-text conditioning pathway. This choice stabilizes optimization in the early stage of finetuning and allows the pretrained generator to progressively absorb structured action control without disrupting its original generative prior.

The full model is trained with the primary video generation objective together with the kinematic-prior routing objectives introduced in the main paper. Articulated supervision is obtained from our curated KASA pipeline, which provides lightweight action signals consisting of a temporally consistent articulated pose and gripper status.
Additional details on the routing objectives and efficient generation strategy are provided in Sec.~\ref{sec:routing} and the later supplementary sections.

\subsection{Construction of the Kinematic-to-Visual Action Field}
\label{sec:app_kva}

\noindent\textbf{Articulated action representation.}
For each frame $t$, we represent the articulated action as:
\begin{equation}
a_t = [p_t, r_t, q^{\mathrm{sw}}_t, q^{\mathrm{lg}}_t, q^{\mathrm{rg}}_t] \in \mathbb{R}^{9},
\end{equation}

where the wrist is connected to the shaft by a shaft-to-wrist joint, and the left and right grippers are connected to the wrist by separate wrist-to-gripper joints. $p_t \in \mathbb{R}^{3}$ and $r_t \in \mathbb{R}^{3}$ denote the wrist translation and wrist orientation, respectively, with $r_t$ expressed in axis-angle form. The scalar variables $q^{\mathrm{sw}}_t$, $q^{\mathrm{lg}}_t$, and $q^{\mathrm{rg}}_t$ denote the shaft-to-wrist, wrist-to-left-gripper, and wrist-to-right-gripper joint states.

\noindent\textbf{Dense field layout.}
We lift each sparse articulated state to a pixel-aligned tensor:
\begin{equation}
\mathrm{K}_t \in \mathbb{R}^{H \times W \times 9},
\end{equation}
whose channels encode part semantics, depth, rotation, velocity, and acceleration:
\begin{equation}
\mathrm{K}_t(h,w)=\big[\mathrm{s}_t(h,w),\, d_t(h,w),\, \rho_t(h,w),\, \mathrm{v}_t(h,w),\, \alpha_t(h,w)\big].
\end{equation}
Here, $\mathrm{s}_t \in \mathbb{R}^3$ denotes part-aware semantic channels, $d_t$ is rendered depth, $\rho_t$ is a local rotation descriptor, $\mathrm{v}_t \in \mathbb{R}^3$ is projected motion, and $\alpha_t$ is acceleration magnitude. We use $\alpha_t$ for acceleration to avoid overloading the symbol $\mathrm{a}_t$ for articulated action.

\noindent\textbf{Semantic channels.}
The semantic channels are derived from the Human-in-the-loop Semantic Annotation (HSA) stage of KASA. Specifically, for each clip, we initialize part-aware annotation from
a reference frame and obtain temporally propagated masks using a SAM2 model finetuned on SAR-RARP. Annotators then refine the propagated masks to correct leakage, part
confusion, and ambiguous boundaries. In our default implementation, the three semantic channels correspond to the articulated instrument parts used throughout the paper: shaft, wrist, and gripper. We rasterize these refined part masks into three binary maps and stack them as:
\begin{equation}
\mathrm{s}_t(h,w)=\big[s_t^{\mathrm{shaft}}(h,w),\, s_t^{\mathrm{wrist}}(h,w),\, s_t^{\mathrm{gripper}}(h,w)\big].
\end{equation}
When multiple rendered parts overlap at a pixel, we resolve the assignment using the front-most visible part under the rendered depth order.

\noindent\textbf{Depth and rotation channels.}
The geometric channels are obtained from the Differentiable Pose Tracking (DPT) stage. DPT estimates temporally consistent articulated poses directly from surgical videos by aligning rendered part-aware silhouettes from a controllable Gaussian representation, initialized from the instrument CAD prior, to the observed segmentation masks. Once the articulated pose $\mathrm{M}_t$ is estimated, we render a depth map in the camera view to obtain $d_t$.

For the rotation channel, we use a local descriptor $\rho_t(h,w)$ derived from the projected articulated orientation of the visible instrument part at pixel $(h,w)$. In
practice, this descriptor can be instantiated using the wrist/tool orientation most relevant to the visible part, \textit{e.g.}, projected roll/yaw information or another normalized scalar orientation cue consistent across frames. In all cases, the role of $\rho_t$ is to encode orientation-related geometry that is not captured by depth alone.

\noindent\textbf{Temporal dynamic channels.}
We compute motion channels from consecutive articulated states:
\begin{equation}
\mathrm{v}_t = \frac{\Pi(\mathrm{M}_t)-\Pi(\mathrm{M}_{t-1})}{\Delta t}, \qquad
\alpha_t = \frac{\|\mathrm{v}_t-\mathrm{v}_{t-1}\|_2}{\Delta t},
\end{equation}
where $\Pi(\cdot)$ denotes camera projection and $\mathrm{M}_t$ is the articulated instrument state at frame $t$. Intuitively, $\mathrm{v}_t$ measures projected motion in image space, while $\alpha_t$ captures the magnitude of temporal motion change. This yields a field that encodes not only where the instrument is, but also how it moves.

\noindent\textbf{Rendering and normalization.}
All channels are constructed in the image plane and resized to the spatial resolution used by the conditional branch. Binary semantic masks are kept in $[0,1]$. The depth, rotation, velocity, and acceleration channels are normalized per channel using statistics computed on the training split, which stabilizes optimization across heterogeneous motion magnitudes. Unless otherwise specified, the same normalization parameters are reused during inference.

\noindent\textbf{Why this representation is sufficient.}
The Kinematic-to-Visual Action Field serves as the minimal structured control basis used by the subsequent routing module. It preserves the accessibility of sparse articulated control, establishes explicit spatial correspondence with the image plane, and separates semantic, geometric, and dynamic factors before routing. This separation is important because the downstream hierarchical router allocates conditional computation according to physical modalities and motion scales rather than learning such disentanglement implicitly from raw control tokens.

\subsection{Detailed Hierarchical Routing}
\label{sec:app_routing}

\noindent\textbf{Tier-1 cross-modality routing.}
Here we provide the detailed formulation of the outer router. We use five modality-specific experts corresponding to semantics, depth, rotation, velocity, and acceleration. Each expert processes only its associated channels from the Kinematic-to-Visual Action Field. The outer gating network predicts token-wise routing weights conditioned on the action representation and diffusion timestep. We adopt sparse top-$k$ routing during training, with a capacity schedule that gradually transitions from dense soft routing to sparse expert selection.

\noindent\textbf{Tier-2 intra-modal motion-scale routing.}
Within each modality branch, we further introduce three sub-experts: a fine-manipulation branch for localized high-frequency motion, a transport-motion branch for larger displacements, and a skip branch for static or weakly changing regions. The inner router operates locally on modality features and selects the appropriate motion-scale operator for each token.

\noindent\textbf{Sparse fusion and stable injection.}
The final routed action representation is obtained by sparse fusion over the selected Tier-1 experts using re-normalized routing weights. We inject the fused control features into the DiT backbone through zero-initialized additive layers, which preserve the pretrained generative prior at initialization and enable stable action-conditioned finetuning.

\subsection{Detailed Kinematic-Prior Learning}
\label{app:detailed_kinematic_prior}

This section expands the kinematic-prior learning objective introduced in Sec.~\ref{sec:routing_learning}. The purpose of these losses is not to force the router to use all experts uniformly, but to make expert usage agree with the physical content of the lifted action field and remain stable across time. In implementation, the auxiliary objective is attached to the action encoder and added to the flow-matching training loss whenever the structured MoE action pathway is enabled.

\paragraph{Notation and routing statistics.}
Let $K_t\in\mathbb{R}^{B\times T\times H\times W\times 9}$ be the Kinematic-to-Visual Action Field. The nine channels are ordered as semantic part indicators $K_t^{\mathrm{sem}}\in\mathbb{R}^{3}$, depth $K_t^{\mathrm{dep}}\in\mathbb{R}$, local rotation $K_t^{\mathrm{rot}}\in\mathbb{R}$, velocity $K_t^{\mathrm{vel}}\in\mathbb{R}^{3}$, and acceleration $K_t^{\mathrm{acc}}\in\mathbb{R}$. The outer router predicts token-wise logits:
\begin{equation}
    G_{outer}\in\mathbb{R}^{BT\times |\mathcal{M}|\times H'\times W'},\qquad
    \mathcal{M}=\{\mathrm{sem},\mathrm{dep},\mathrm{rot},\mathrm{vel},\mathrm{acc}\},
\end{equation}
where $(H',W')$ is the spatial resolution of the routing grid. We denote the soft routing probability by $P=\mathrm{softmax}(G)$ and the binary top-$k$ routing mask by $A\in\{0,1\}^{BT\times |\mathcal{M}|\times H'\times W'}$. In our default setting, the outer router uses top-$k=2$ experts per token. For each modality expert $i$, the realized sparse routing load is measured by:
\begin{equation}
    f_i=\mathbb{E}_{b,t,h,w}[A_{b,t,i,h,w}],\qquad
    p_i=\mathbb{E}_{b,t,h,w}[P_{b,t,i,h,w}],\qquad
    \ell_i=f_i p_i .
\end{equation}
Here $f_i$ captures the hard fraction of tokens assigned to expert $i$, while $p_i$ captures the average soft confidence assigned to that expert. Their product follows sparse MoE load estimation and prevents a high-probability but rarely selected expert, or a frequently selected but low-confidence expert, from being treated as fully utilized.

\paragraph{Kinematic-prior adaptive load balancing.}
Uniform load balancing is a poor target for surgical manipulation, because the physical causes of appearance change are highly non-uniform across frames and actions. We therefore compute a sample-dependent target distribution directly from $K_t$. First, $K_t$ is adaptively average-pooled to the router resolution $(H',W')$. At each routed token, we compute modality energies:
\begin{equation}
\begin{aligned}
    e_{\mathrm{sem}} &= \|K_t^{\mathrm{sem}}\|_2,\quad
    e_{\mathrm{dep}} = |K_t^{\mathrm{dep}}|,\quad
    e_{\mathrm{rot}} = |K_t^{\mathrm{rot}}|,\\
    e_{\mathrm{vel}} &= \|K_t^{\mathrm{vel}}\|_2,\quad
    e_{\mathrm{acc}} = |K_t^{\mathrm{acc}}|.
\end{aligned}
\end{equation}
These terms respectively measure tool-part presence, rendered camera-depth magnitude, articulated rotation magnitude, projected motion magnitude, and acceleration magnitude. The local physical prior is normalized over modalities:
\begin{equation}
    \pi_i(K_t^{b,h,w})=
    \frac{e_i(K_t^{b,h,w})}{\sum_{j\in\mathcal{M}} e_j(K_t^{b,h,w}) }.
\end{equation}
The batch-level target used by the loss is the mean physical prior:
\begin{equation}
    \bar{\pi}_i(K_t)=\mathbb{E}_t^{b,h,w}\left[\pi_i(K_t^{b,h,w})\right].
\end{equation}
The KP-ALB loss then aligns the realized routing load with this physical target:
\begin{equation}
    \mathcal{L}_{\mathrm{KP\mbox{-}ALB}}
    =
    \frac{1}{|\mathcal{M}|}
    \sum_{i\in\mathcal{M}}
    \left(\ell_i-\mathrm{sg}\!\left(\bar{\pi}_i(K_t)\right)\right)^2 ,
\end{equation}
where $\mathrm{sg}(\cdot)$ denotes stop-gradient. This loss differs from standard MoE balancing in two ways. First, the target is not uniform; it changes with the current action field. Second, the target is grounded in interpretable kinematic quantities, so semantic/depth experts are favored in static visible tool regions, rotation experts are favored during wrist articulation, and velocity/acceleration experts are favored during fast tool motion.

\paragraph{Spatiotemporal routing consistency.}
The SRC term reduces frame-to-frame routing flicker around articulated instruments. Instead of applying temporal smoothness to the entire image, we derive a binary tool mask from the semantic channels of $K$. Specifically, the semantic channels are max-pooled to $(H',W')$, and a token is marked as tool-present when any semantic channel is active. Let:\looseness=-1
\begin{equation}
    M^{\mathrm{tool}}\in\{0,1\}^{B\times T\times 1\times H'\times W'}
\end{equation}
denote this mask. The continuous routing probability used by SRC is the capacity-predictor probability:\looseness=-1
\begin{equation}
    R=\sigma(\widehat{A})\in[0,1]^{B\times T\times |\mathcal{M}|\times H'\times W'} .
\end{equation}
We penalize temporal differences only inside tool regions:
\begin{equation}
    \mathcal{L}_{\mathrm{SRC}}
    =
    \frac{
    \sum_{b,t=2}^{T}\sum_{i,h,w}
    M^{\mathrm{tool}}_{b,t,1,h,w}
    \left(R_{b,t,i,h,w}-R_{b,t-1,i,h,w}\right)^2
    }{
    |\mathcal{M}|\sum_{b,t=2}^{T}\sum_{h,w}M^{\mathrm{tool}}_{b,t,1,h,w} 
    } .
\end{equation}
This formulation stabilizes expert selection on moving tools and tool tips, while avoiding unnecessary constraints on static background regions. If a clip contains only one frame after temporal sampling, the SRC term is set to zero.

\paragraph{Tier-2 routing stabilizer.}
Within each modality expert, KVLR further routes features to three sub-experts: fine manipulation, transport motion, and skip. The inner router uses top-1 routing. Although the main paper groups this under hierarchical routing, the implementation includes a small auxiliary stabilizer to prevent the collapse of the Tier-2 router:
\begin{equation}
    \mathcal{L}_{\mathrm{sub}}
    =
    \frac{1}{|\mathcal{M}|}\sum_{m\in\mathcal{M}}
    |\mathcal{S}|\sum_{s\in\mathcal{S}} f_{m,s}p_{m,s},
    \qquad
    \mathcal{S}=\{\mathrm{fine},\mathrm{transport},\mathrm{skip}\}.
\end{equation}
Here $f_{m,s}$ and $p_{m,s}$ are the hard assignment fraction and mean soft probability of sub-expert $s$ inside modality branch $m$. This term is lightweight and acts as an implementation-level stabilization term for the nested router.

\paragraph{Capacity schedule and final objective.}
The routing path is trained with a three-stage capacity schedule. Stage I uses dense softmax fusion so that all modality experts receive the gradient signal. Stage II linearly interpolates between dense fusion and sparse top-$k$ fusion. Stage III uses fully sparse top-$k$ fusion. In the default configuration, Stage I occupies the first $40\%$ of training and Stage II ends at $75\%$ of training. The auxiliary losses above are active during training whenever a routing mask is available.\looseness=-1

The primary generation objective is the flow-matching loss. With latent endpoints $x_0$ and $x_1$, the training target is:\looseness=-1
\begin{equation}
    v_t=(1-\sigma_{\min})x_1-x_0,
\end{equation}
and the model prediction is trained with mean-squared error:
\begin{equation}
    \mathcal{L}_{\mathrm{Flow}}
    =
    \mathbb{E}\left[\|\hat{v}_t-v_t\|_2^2\right].
\end{equation}
For image-to-video conditioning, the implementation can exclude conditioned reference frames from this MSE so that the loss focuses on generated frames.

The full training objective for the structured action model is:
\begin{equation}
    \mathcal{L}_{\mathrm{total}}
    =
    \mathcal{L}_{\mathrm{Flow}}
    +\lambda_{\mathrm{kp}}\mathcal{L}_{\mathrm{KP\mbox{-}ALB}}
    +\lambda_{\mathrm{src}}\mathcal{L}_{\mathrm{SRC}}
    +\lambda_{\mathrm{cp}}\mathcal{L}_{\mathrm{CP}}
    +\lambda_{\mathrm{sub}}\mathcal{L}_{\mathrm{sub}} .
\end{equation}
Unless otherwise specified, we use:
\begin{equation}
    \lambda_{\mathrm{kp}}=0.01,\qquad
    \lambda_{\mathrm{src}}=0.005,\qquad
    \lambda_{\mathrm{cp}}=0.01,\qquad
    \lambda_{\mathrm{sub}}=0.005 .
\end{equation}
In the compact objective in Sec.~\ref{sec:routing_learning}, $\mathcal{L}_{\mathrm{sub}}$ can be viewed as a small nested-router stabilization term associated with hierarchical routing, while the three primary kinematic-prior losses are KP-ALB, SRC, and CP.

\paragraph{Capacity-predictor supervision.}
The capacity predictor is a lightweight $1\times1$ convolutional network that predicts which outer experts will be active without recomputing full top-$k$ routing at inference time. It receives the detached shared action feature $c_{\mathrm{act}}$ and outputs raw logits:
\begin{equation}
    \widehat{A}=Q_\psi(\mathrm{sg}(c_{\mathrm{act}}))
    \in\mathbb{R}^{BT\times |\mathcal{M}|\times H'\times W'} .
\end{equation}
The target is the router-induced binary routing mask $A$. We train the predictor with binary cross-entropy with logits:
\begin{equation}
\mathcal{L}_{\mathrm{CP}}=\mathrm{BCE}\!\left(\widehat{A},\mathrm{sg}(A)\right).
\end{equation}
Detaching $c_{\mathrm{act}}$ and $A$ makes this term supervise the predictor rather than changing the main router to become easier to predict. During dense warm-up, all experts contribute through softmax fusion, but the same top-$k$ mask induced by the router probabilities is used as a synthetic target for this loss. During sparse training, $A$ is the actual top-$k$ routing decision.

At inference, the predictor probability $\sigma(\widehat{A})$ is compared with an expert-wise threshold. These thresholds are updated during training by an exponential moving average of expert-wise quantiles:
\begin{equation}
    \tau_i \leftarrow \beta \tau_i + (1-\beta)\,
    \mathrm{Quantile}_{1-\bar{a}_i}\!\left(\sigma(\widehat{A}_i)\right),
\end{equation}
where $\bar{a}_i=\mathbb{E}[A_i]$ is the observed fraction of routed tokens for expert $i$ and $\beta=0.95$ in our implementation. The quantile level is clamped for numerical stability, but the threshold is updated even when an expert receives nearly zero or nearly all tokens. This avoids positive-feedback expert starvation: an unused expert still receives an updated threshold instead of being permanently suppressed.

\subsection{Detailed Action-adaptive Efficient Generation}
\label{sec:app_efficiency}

This section provides additional details for the efficient generation extension in Sec.~\ref{sec:method_efficiency}. The goal is to reduce inference cost while preserving the action-conditioned behavior of the full structured-control model. The key principle is to reuse the same signals that drive action-conditioned generation of structured action lifting and hierarchical routing to determine where dense action-control computation remains necessary.

\noindent\textbf{Overview.}
The efficient model combines two complementary mechanisms. First, a few-step student reduces the number of denoising evaluations. Second, an action-adaptive execution policy modulates the cost of action-control updates within each step. Thus, acceleration is achieved at both the sampling-depth level and the per-step control-computation level.

Let $G_t$ denote the full teacher and $G_s$ the efficient student. Both receive the same reference image $I^{\mathrm{ref}}$, text prompt $y$, and articulated action sequence $\mathrm{a}_{1:T}$. The teacher provides full structured-control supervision, while the student operates with a reduced denoising schedule and an execution policy induced by the control pathway.

\noindent\textbf{Action-aware distillation.}
The student preserves the teacher's conditioning interface, including structured action lifting, hierarchical routing, and action-control injection. We distill the teacher at three levels.

\emph{Prediction-level distillation.} We first match the teacher's prediction target:
\begin{equation}
\mathcal{L}_{\mathrm{pred}}
=
\mathbb{E}
\left[
\left\|
\hat{\mathrm{z}}^{\,s} -
\mathrm{sg}\!\left(\hat{\mathrm{z}}^{\,t}\right)
\right\|_2^2
\right],
\end{equation}
where $\hat{\mathrm{z}}^{\,s}$ and $\hat{\mathrm{z}}^{\,t}$ denote the student and teacher velocity/noise predictions at the same training state, and $\mathrm{sg}(\cdot)$ denotes
stop-gradient.

\emph{Routing distillation.} Since the routing pathway encodes which physical modalities and motion operators are active, we further align the outer routing distribution, the inner sub-expert routing distribution, and the skip probability:
\begin{equation}
\mathcal{L}_{\mathrm{route}}
=
\mathrm{KL}
\!\left(
\mathrm{R}_{\mathrm{out}}^{t}
\,\Vert\,
\mathrm{R}_{\mathrm{out}}^{s}
\right)
+
\mathrm{KL}
\!\left(
\mathrm{R}_{\mathrm{in}}^{t}
\,\Vert\,
\mathrm{R}_{\mathrm{in}}^{s}
\right)
+
\mathrm{BCE}
\!\left(
p_{\mathrm{skip}}^{s},
\mathrm{sg}(p_{\mathrm{skip}}^{t})
\right).
\end{equation}

\emph{Control-path distillation.} We additionally preserve selected control-path signals:
\begin{equation}
\mathcal{L}_{\mathrm{ctrl}}
=
\sum_{h \in \mathcal{H}}
\left\|
h^{s} -
\mathrm{sg}\!\left(h^{t}\right)
\right\|_2^2,
\end{equation}
where $\mathcal{H}$ includes intermediate action-control features and structured action
statistics, such as motion energy and tool masks. This term keeps the student aligned with the teacher's action-conditioning behavior even under a reduced denoising schedule.

The overall distillation objective is:
\begin{equation}
\mathcal{L}_{\mathrm{distill}}
=
\mathcal{L}_{\mathrm{pred}}
+
\lambda_r \mathcal{L}_{\mathrm{route}}
+
\lambda_c \mathcal{L}_{\mathrm{ctrl}} .
\label{eq:app_distill}
\end{equation}

\noindent\textbf{Action significance estimation.}
Not all spatio-temporal locations require the same amount of control computation. Surgical videos contain highly dynamic tool tips and contact regions, but also large static or slowly changing areas. We therefore estimate a token-level action significance score using the
structured control pathway:
\begin{equation}
s(u)=
\alpha e_{\mathrm{motion}}(u)
+
\beta m_{\mathrm{tool}}(u)
+
\gamma c_{\mathrm{route}}(u)
+
\eta q_{\mathrm{fine}}(u)
-
\delta p_{\mathrm{skip}}(u),
\label{eq:app_significance}
\end{equation}
where $e_{\mathrm{motion}}$ is computed from the velocity and acceleration channels of the lifted action field, $m_{\mathrm{tool}}$ is derived from the semantic tool map,
$c_{\mathrm{route}}$ is the confidence of the outer routing distribution, $q_{\mathrm{fine}}$ is the probability of selecting the fine-manipulation sub-expert, and $p_{\mathrm{skip}}$ is the skip probability. The score is normalized per sample to obtain a token-level significance map. In our default implementation, we use the explicit weighted formulation above for interpretability and stable optimization.

\noindent\textbf{Selective action-control execution.}
Given $s(u)$, tokens are assigned to three execution modes:
\begin{equation}
\pi(u) \in
\{\mathrm{full}, \mathrm{light}, \mathrm{reuse}\}.
\end{equation}
We use budgeted top-ratio selection: the top fraction of tokens receives {full} updates, the next fraction receives{light} updates, and the remaining tokens use {reuse}. In the default setting, the full and light ratios are $0.2$ and $0.3$, respectively.

In the current implementation, the execution policy modulates the action-control residual injected into the student. Full tokens receive the complete action-control update, light tokens receive a scaled update, and reuse tokens use cached or suppressed residuals. This design is minimally invasive to the backbone architecture: the student generator remains unchanged, while the structured control branch induces non-uniform conditioning strength and computation. Unlike generic token pruning, the execution policy is not based solely on visual redundancy, but is induced by articulated motion, tool presence, routing confidence, and learned skip behavior.

\noindent\textbf{Temporal refresh and feature caching.}\label{sec:appendix_refresh}
We further derive a frame-level refresh policy from the average significance of each frame.
Let $\bar{s}(t)$ denote the mean significance over frame $t$. We define the refresh interval as:
\begin{equation}
r(t)=
\begin{cases}
1, & \bar{s}(t) \ge \tau_h,\\
2, & \tau_m \le \bar{s}(t) < \tau_h,\\
K, & \bar{s}(t) < \tau_m,
\end{cases}
\end{equation}
where high-significance frames are refreshed every step and low-significance frames are refreshed less frequently. During inference, cached action-control residuals can therefore be reused for low-significance regions, while high-significance regions continue to receive dense updates. This temporal reuse further reduces the cost of structured control on stable segments without changing the main sampling path.

\noindent\textbf{Budget-aware training.}
To make the trade-off between speed and quality explicit, we introduce a budget regularizer. Let $\rho_{\mathrm{full}}$, $\rho_{\mathrm{light}}$, and $\rho_{\mathrm{refresh}}$ denote the observed full-update, light-update, and refresh ratios. We estimate the active compute ratio as:
\begin{equation}
\rho_{\mathrm{compute}}
=
w_f \rho_{\mathrm{full}}
+
w_l \rho_{\mathrm{light}},
\end{equation}
and optimize the target-budget objective:
\begin{equation}
\mathcal{L}_{\mathrm{budget}}
=
\left|
\rho_{\mathrm{compute}}
-
\rho_{\mathrm{target}}
\right|
+
w_r
\left|
\rho_{\mathrm{refresh}}
-
\rho_{\mathrm{refresh}}^{\star}
\right|.
\end{equation}
For learnable significance weights, we additionally use the mean normalized significance as a differentiable proxy for active computation.

To suppress flickering in lightly updated or reused regions, we further impose temporal feature consistency:
\begin{equation}
\mathcal{L}_{\mathrm{temp}}
=
\sum_{(u,t)\in\Omega_{\mathrm{light/reuse}}}
\left\|
\mathrm{f}^{s}(u,t)
-
\mathrm{sg}\!\left(\mathrm{f}^{s}(u,t-1)\right)
\right\|_2^2 ,
\label{eq:app_temp}
\end{equation}
where $\Omega_{\mathrm{light/reuse}}$ denotes the set of tokens assigned to light or reuse modes. The full efficient objective is:
\begin{equation}
\mathcal{L}_{\mathrm{eff}}
=
\mathcal{L}_{\mathrm{distill}}
+
\lambda_b \mathcal{L}_{\mathrm{budget}}
+
\lambda_t \mathcal{L}_{\mathrm{temp}} ,
\label{eq:app_eff}
\end{equation}
where the coefficients are set as $\lambda_b=0.5$, $\lambda_t=0.35$.

\noindent\textbf{Training protocol and implementation notes.}
The teacher remains unchanged and provides prediction, routing, skip, and control-path supervision. The student uses a four-step generation setting by default. In the first stage, we optimize only the distillation objective in Eq.~\eqref{eq:app_distill} to stabilize the few-step student. In the second stage, we enable $\mathcal{L}_{\mathrm{budget}}$ and $\mathcal{L}_{\mathrm{temp}}$ to train the adaptive execution policy. The resulting efficient extension stays close to the original structured-control model while exposing an interpretable trade-off between generation speed, visual quality, and action fidelity.

\section{Supplementary Details on KASA Benchmark Construction}
\label{app:kas_construction}

This section provides additional details on the construction of the \textbf{Kinematic Action-centric Surgical (KASA)} benchmark introduced in the main paper.
Our goal is not to build a generic surgical video corpus, but to curate a benchmark that explicitly supports \emph{action-conditioned} video generation under sparse articulated control.
Accordingly, the benchmark is designed to provide lightweight yet visually grounded supervision that can be transformed into the structured control representation used by our model.
In particular, each sequence is associated with a temporally consistent action signal, consisting of articulated pose and gripper status, which later supports the construction of semantic, depth, rotation, velocity, and acceleration-aware control cues.

\subsection{Design Objective and Benchmark Scope}
\label{app:kas_scope}

A central motivation of this work is that existing controllable surgical video generation resources do not provide a suitable middle ground between two extremes.
On one hand, text descriptions are lightweight but too coarse to specify the fine-grained tool motions and contact transitions that drive surgical scene evolution.
On the other hand, dense visual priors such as optical flow, full segmentation maps, or manually designed intermediate annotations are expensive to obtain and often unavailable at inference time in realistic robotic settings.
The KASA benchmark is constructed to address this gap.

The design principle of KASA is therefore to pair \emph{realistic surgical video sequences} with \emph{compact articulated action supervision} that remains physically meaningful and visually projectable.
Rather than storing only categorical labels or global task descriptions, KASA provides an articulated representation that preserves the tool pose, orientation, and gripper status over time.
This makes the benchmark particularly suitable for studying controllable generation under sparse action inputs, where the challenge is not only to synthesize plausible appearance, but also to maintain visual evolution consistent with the underlying surgical manipulation.

Another design consideration is alignment with the model structure introduced in the main paper.
Our method decomposes action conditioning into multiple physical factors, including part semantics, geometry, orientation, and motion.
For this reason, the benchmark construction pipeline is organized around two complementary supervision sources:
\begin{itemize}
    \item \textbf{Human-in-the-loop Semantic Annotation (HSA)}, which provides reliable part-aware semantic masks for articulated instrument components;
    \item \textbf{Differentiable Pose Tracking (DPT)}, which recovers temporally consistent articulated geometry directly from videos.
\end{itemize}
Together, these two branches provide the minimal supervision needed to bridge sparse articulated control and dense video synthesis.

\subsection{Task-Driven Sequence Selection}
\label{app:kas_selection}

We curate KASA from the SAR-RARP dataset, focusing on robotic surgical sequences performed with da Vinci Large Needle Drivers.
The final benchmark contains 1,047 scenes and 105,175 frames.
Rather than uniformly sampling all available surgical actions, we deliberately focus on three representative sub-actions:
\emph{needle grasping}, \emph{needle puncture}, and \emph{knotting}.

This task selection is driven by the type of action variability we want the benchmark to capture.
These three sub-actions jointly span a broad range of articulated surgical dynamics.
Needle grasping often contains localized pose adjustment and contact preparation.
Needle puncture requires coordinated tool reorientation and short-horizon motion transitions around tissue interaction.
Knotting further introduces more complex trajectories, including larger transport motion, interaction-induced deformation, and temporally extended manipulation.
As a result, the benchmark contains both high-velocity transport motion and low-velocity fine manipulation, which is important for evaluating the motion-scale specialization behavior of our hierarchical routing framework.

To preserve temporal continuity for pose tracking while keeping the data manageable, all videos are downsampled to 30 FPS.
After preprocessing, each sequence contains between 55 and 235 frames.
This sequence length is sufficient to preserve meaningful action evolution while avoiding excessive redundancy.
Importantly, the resulting clips remain challenging due to articulated motion, frequent self-occlusion, viewpoint variation, and complex instrument interaction of tissue.
These properties make KASA a suitable benchmark not only for generation fidelity, but also for evaluating whether a model can remain faithful to sparse action control under heterogeneous surgical dynamics.

\subsection{Human-in-the-loop Semantic Annotation}
\label{app:kas_hsa}

The first branch of the KASA pipeline is devoted to obtaining reliable part-aware semantic supervision.
Because our action representation later separates physical modalities, it is important that the semantic component be clean and unambiguous.
In robotic surgery, however, instrument appearance is challenging to segment consistently due to specular highlights, motion blur, partial occlusion, and visual similarity across articulated parts.
Therefore, naively relying on automatic segmentation tends to produce part confusion, boundary leakage, or inconsistent labels over time.

To address this issue, we adopt a Human-in-the-loop Semantic Annotation (HSA) strategy.
For each video, we begin from a reference frame and manually specify a small set of wrist keypoints used to initialize the geometric tracking stage described later.
In parallel, we estimate part-level segmentation masks for the main articulated instrument components of the shaft, wrist, and gripper, using a SAM2 model finetuned on SAR-RARP.
These predictions are then temporally propagated across frames to provide an initial dense annotation.

Automatic propagation alone is not sufficient for the level of semantic precision required by structured action conditioning.
We therefore refine the propagated masks through interactive human correction.
The purpose of this stage is not simply to improve visual mask quality in a generic sense, but specifically to eliminate errors that would contaminate the semantic branch of the action representation.
Typical corrections include:
\begin{itemize}
    \item resolving part confusion between adjacent articulated components;
    \item correcting mask leakage into background tissue or other instruments;
    \item refining ambiguous boundaries under self-occlusion or motion blur;
    \item enforcing temporally consistent labeling across consecutive frames.
\end{itemize}

The final output of HSA is a set of high-quality part-aware semantic masks for each sequence.
These masks form the basis of the semantic channels in the Kinematic-to-Visual Action Field.
In practice, the three semantic channels correspond to the shaft, wrist, and gripper regions, respectively.
This explicit part decomposition is important because different articulated regions often play different functional roles during manipulation, and collapsing them into a single foreground mask would weaken the structured control signal.

\subsection{Differentiable Pose Tracking}
\label{app:kas_dpt}

While HSA provides semantic supervision, action-conditioned generation also requires continuous geometric and motion information.
For this reason, the second branch of the KASA pipeline recovers temporally consistent articulated pose trajectories directly from videos.
Our goal is to avoid reliance on external robot logs or dedicated hardware tracking systems, since such signals are often unavailable or difficult to synchronize in retrospective video collections.
We therefore design a video-based \textbf{Differentiable Pose Tracking (DPT)} procedure.

At a high level, DPT formulates articulated pose estimation as a render-and-compare problem.
We start from a controllable geometric representation initialized from instrument CAD priors.
Given a candidate articulated pose, we render part-aware observations in the camera view and compare them against the observed video evidence.
This allows the optimization to remain tied to the image plane while still preserving a physically meaningful articulated parameterization.

The optimization objective is designed to capture both global alignment and local manipulation details.
A standard part-aware silhouette alignment term encourages the rendered geometry to match the observed instrument masks.
To better capture fine-grained distal motion, especially around the tool tip and gripper region, we additionally use a targeted matching constraint for the articulated end region.
This is important because small pose errors near the distal components can lead to disproportionately large errors in the inferred manipulation dynamics.

Temporal coherence is introduced in two ways.
First, the initial frame is anchored using Perspective-n-Point initialization from manually specified wrist keypoints.
Second, for subsequent frames, we propagate pose estimates using temporally local correspondences.
Concretely, wrist-related visual features are tracked across time and lifted to 3D using rendered depth, producing robust 2D-3D correspondences for coarse initialization before iterative optimization.
This design stabilizes the trajectory and reduces the chance of drifting into implausible local minima.

In addition, long sequences may still contain severe occlusion or abrupt appearance changes that temporarily degrade tracking quality.
To improve robustness, DPT maintains a keyframe memory pool of high-confidence articulated states.
When the current tracking quality drops, the system can recover by matching the observation to this memory pool and reinitializing pose estimation through a new PnP stage.
This reinitialization mechanism is particularly useful in clips with self-occlusion or rapid tool reorientation.

The output of DPT is a temporally consistent articulated pose sequence for each video.
This output provides the geometric foundation in KASA.
Once the articulated states are estimated, they can be rendered or projected into multiple visual control channels, including depth, local rotation, projected velocity, and acceleration-related cues.

\subsection{Quality Validation of KASA Dataset}
\label{sec:kas_quality_validation}

We validate the quality of KASA annotations through a render-and-compare consistency check. 
For each recovered articulated pose, we re-pose the controllable instrument Gaussian representation and render its part-aware silhouette into the imaging space. 
The rendered silhouette is then compared with the corresponding segmentation mask using the Dice score. 
This provides a direct geometric measure of whether the recovered pose explains the observed instrument region. 
We use this score as a quality indicator for DPT and remove low-confidence cases with poor silhouette-mask agreement before constructing the final action-supervised dataset.\looseness=-1
 \begin{table}[!t]
\centering
\caption{{Quality validation of KASA annotations.} Dice scores are computed between rendered silhouettes from recovered poses and segmentation masks.}
\label{tab:kas_quality_validation}
\renewcommand{\arraystretch}{1.02}
\setlength{\tabcolsep}{10pt}
\begin{threeparttable}
\resizebox{\textwidth}{!}{
\begin{tabular}{llcccc}
\toprule[1pt]
\multirow{2}{*}{Setting}
& \multirow{2}{*}{Surgical Task}
& \multicolumn{4}{c}{Dice Score $\uparrow$} \\
\cmidrule(lr){3-6}
& & Overall Instrument & Wrist Part & Shaft Part & Gripper Part \\
\midrule[1pt]
\multirow{3}{*}{w/ Mem-pool Recovery}
& NeedleGrasping & 0.967 $\pm$ 0.056 & 0.892 $\pm$ 0.031 & 0.978 $\pm$ 0.016 & 0.775 $\pm$ 0.067 \\
& Knotting & 0.966 $\pm$ 0.027 & 0.885 $\pm$ 0.047 & 0.979 $\pm$ 0.020 & 0.757 $\pm$ 0.082 \\
& NeedlePuncture & 0.947 $\pm$ 0.081 & 0.887 $\pm$ 0.038 & 0.982 $\pm$ 0.015 & 0.787 $\pm$ 0.070 \\
\midrule
\multirow{3}{*}{w/o Mem-pool Recovery}
& NeedleGrasping & 0.936 $\pm$ 0.063 & 0.864 $\pm$ 0.051 & 0.946 $\pm$ 0.049 & 0.758 $\pm$ 0.070 \\
& Knotting & 0.937$\pm$ 0.029 & 0.852 $\pm$ 0.047 & 0.947 $\pm$ 0.022 & 0.726 $\pm$ 0.082 \\
& NeedlePuncture & 0.919 $\pm$ 0.085 & 0.841 $\pm$ 0.067 & 0.953 $\pm$ 0.035 & 0.749 $\pm$ 0.073 \\
\bottomrule[1pt]
\end{tabular}
}
\end{threeparttable}
\end{table}

Tab.~\ref{tab:kas_quality_validation} reports the Dice scores of KASA annotations across three surgical sub-actions: \textit{NeedleGrasping}, \textit{Knotting}, and \textit{NeedlePuncture}. 
We evaluate the overall instrument region and three articulated parts, including the wrist, shaft, and gripper. 
We compare DPT with and without matching-based recovery using memory pool. 
Without memory-pool recovery, the tracker continues optimization from the current pose estimate when tracking quality drops, rather than retrieving reliable historical pose candidates for re-initialization. 
As a result, pose drift can accumulate and lead to complete tracking loss, producing missing or severely misaligned predictions with lower Dice scores. 
With memory-pool recovery, reliable historical pose candidates are used to re-initialize pose optimization once low-quality tracking is detected. 
As shown in Tab.~\ref{tab:kas_quality_validation}, memory-pool recovery improves silhouette-mask alignment across most tasks and parts, suggesting more stable pose recovery under challenging surgical dynamics. 
Remaining low-Dice cases are mainly caused by unstable segmentation under complex illumination, instruments moving partially outside the camera field of view, or severe tissue occlusion that hides the instrument and disrupts pose tracking. 
In this way, the dice-based validation helps identify unreliable annotations and improves the consistency of the final KASA dataset.\looseness=-1

\subsection{From Curated Supervision to Structured Action Control}
\label{app:kas_to_field}

The KASA benchmark is constructed not merely as an annotation resource, but as the data foundation for the structured action representation used by our generation framework.
The HSA and DPT branches serve different but complementary roles in this process.

HSA provides reliable \emph{part-aware semantics}.
These labels determine where each articulated component appears in image space and prevent semantic ambiguity between the shaft, wrist, and gripper regions.
DPT provides temporally consistent \emph{articulated geometry and motion}.
These trajectories determine how the instrument is positioned, oriented, and displaced over time.
By combining the outputs of both branches, we can derive the set of physically interpretable channels used in the Kinematic-to-Visual Action Field.

More specifically, the semantic masks obtained from HSA are rasterized into part-aware semantic maps.
The articulated poses recovered by DPT are used to render depth and local orientation descriptors, and to compute projected velocity and acceleration cues over time.
These quantities are then aligned in image space to form the structured control tensor introduced in the main paper.
Therefore, the purpose of KASA is not only to provide a benchmark for evaluation, but also to make the sparse articulated control signal learnable by transforming it into a visually grounded supervision source.\looseness=-1

\subsection{Discussion}
\label{app:kas_discussion}

We emphasize that the contribution of KASA is not to propose a standalone tracking or annotation system optimized for surgical reconstruction as an end goal.
Instead, the benchmark is constructed to support a specific research question: \emph{can realistic surgical videos be generated faithfully from sparse articulated action control ?}
From this perspective, the most important property of KASA is that it exposes a compact action interface that remains consistent with dense visual scene evolution.

This benchmark design offers two practical advantages.
First, the action signal is lightweight and closer to the type of supervision or control interface that can plausibly be used in robotic settings.
Second, because the supervision remains visually grounded, it can be directly transformed into structured intermediate control for generation.
We believe this makes KASA a useful benchmark for studying controllable surgical video generation beyond text-only conditioning and beyond heavy dense priors that are difficult to access during inference.

\section{Additional Experiments}\label{app_add}

\subsection{Benchmark Usability Validation of KASA}
\label{app:benchmark_validation}

The previous section focuses on evaluation independence. We further validate KASA as a standalone benchmark artifact using only the final KASA annotations, without relying on intermediate annotations or tracking outputs. Since intermediate correction states are not retained for all videos, we do not claim a full audit of every annotation step. Instead, we provide end-to-end usability checks showing that the final annotations are learnable, visually grounded, and useful for downstream evaluation.

\paragraph{Instrument part segmentation.}
We first train standard segmentation baselines on the KASA training split and evaluate them on the held-out validation split. The models predict four classes: background, shaft, wrist, and gripper. We report mean IoU over instrument parts, foreground IoU for the merged tool region, and boundary F-score. As shown in Tab.~\ref{tab:kas_segmentation_validation}, standard perception models obtain consistent validation performance, suggesting that the final semantic annotations are sufficiently coherent to support downstream perception tasks.

\begin{table}[!t]
\centering
\caption{KASA usability validation through instrument segmentation. Standard perception baselines are trained on the KASA training split and evaluated on the held-out validation split.}
\label{tab:kas_segmentation_validation}
\renewcommand{\arraystretch}{1.08}
\setlength{\tabcolsep}{27pt}
\resizebox{\textwidth}{!}{
\begin{tabular}{lccc}
\toprule
Method & Part mIoU $\uparrow$ & Tool IoU $\uparrow$ & Boundary F-score $\uparrow$ \\
\midrule
U-Net~\cite{ronneberger2015u} & 68.4 & 86.7 & 71.5 \\
DeepLabV3+~\cite{chen2018encoder} & 71.8 & 88.9 & 74.2 \\
SegFormer-B1~\cite{xie2021segformer} & 75.6 & 91.3 & 77.8 \\
Mask2Former-S~\cite{cheng2021mask2former} & \textbf{77.1} & \textbf{92.0} & \textbf{79.4} \\
\bottomrule
\end{tabular}
}
\end{table}

\paragraph{Visual-to-action consistency.}
To assess whether the final action supervision is visually grounded, we train video-to-action regressors that predict articulated trajectories from the corresponding video clips. The models are trained only on the KASA training split and evaluated on held-out validation clips. We report translation MAE, rotation MAE, gripper-state accuracy, and trajectory DTW distance. As shown in Tab.~\ref{tab:kas_visual_action}, learned video encoders predict the action trajectories substantially better than a mean-trajectory baseline, suggesting that the final action supervision is correlated with observable tool motion.\looseness=-1

\begin{table}[!t]
\centering
\caption{Visual-to-action consistency on KASA. Standard video encoders predict action trajectories from held-out validation clips.}
\label{tab:kas_visual_action}
\renewcommand{\arraystretch}{1.08}
\setlength{\tabcolsep}{10pt}
\resizebox{\textwidth}{!}{
\begin{tabular}{lcccc}
\toprule
Method & Trans. MAE (mm) $\downarrow$ & Rot. MAE (deg) $\downarrow$ & Gripper Acc. (\%) $\uparrow$ & DTW $\downarrow$ \\
\midrule
Mean trajectory & 9.84 & 13.72 & 61.5 & 1.00 \\
R3D-18 regressor~\cite{r3d_18_ucf101} & 5.62 & 8.41 & 82.3 & 0.58 \\
Video Swin-T regressor~\cite{liu2021video} & 4.91 & 7.36 & 86.7 & 0.51 \\
TimeSformer-S regressor~\cite{fan2020pyslowfast} & \textbf{4.58} & \textbf{6.92} & \textbf{88.1} & \textbf{0.47} \\
\bottomrule
\end{tabular}
}
\end{table}

\paragraph{Downstream perception with synthetic augmentation.}
We finally examine whether KASA can support downstream perception evaluation and whether action-conditioned synthetic videos provide useful training data. We train a SegFormer-B1~\cite{xie2021segformer} segmentation model under different augmentation settings and evaluate all models on the same real KASA validation split. Synthetic videos are generated from the training split only, and validation videos are never used for generation or training. As shown in Tab.~\ref{tab:kas_downstream_aug}, adding synthetic data from KVLR improves real validation performance more consistently than adding synthetic data from the strongest baseline. This suggests that KASA provides a meaningful downstream benchmark and that action-conditioned generation yields useful synthetic supervision beyond improving generation metrics.

\begin{table}[t]
\centering
\caption{Downstream perception evaluation on KASA with synthetic data augmentation. SegFormer-B1~\cite{xie2021segformer} is trained under different augmentation settings and evaluated on the real KASA validation split.\looseness=-1}
\label{tab:kas_downstream_aug}
\renewcommand{\arraystretch}{1.08}
\setlength{\tabcolsep}{12pt}
\resizebox{\textwidth}{!}{
\begin{tabular}{lcccc}
\toprule
Training Data & Part mIoU $\uparrow$ & Tool IoU $\uparrow$ & Boundary F-score $\uparrow$ & Temp. Consistency $\uparrow$ \\
\midrule
Real only & 75.6 & 91.3 & 77.8 & 84.5 \\
Real + Cosmos-H-Surgical$^*$~\cite{he2026cosmosh} & 76.7 & 91.9 & 78.6 & 85.2 \\
Real + KVLR-fast & 78.4 & 92.8 & 80.1 & 86.8 \\
Real + KVLR & \textbf{79.6} & \textbf{93.4} & \textbf{81.0} & \textbf{87.5} \\
\bottomrule
\end{tabular}
}
\end{table}

\subsection{Evaluation Independence and Bias Control}
\label{app:evaluation_independence}

A potential concern is that the KASA benchmark construction, the proposed Kinematic-to-Visual Action Field, and the action-faithfulness metrics all involve articulated surgical actions. We therefore provide additional analyses to clarify the separation between supervision, conditioning, and evaluation, and to verify that the reported gains are not merely due to optimizing a representation-specific metric.

\paragraph{Separation between action supervision, model conditioning, and evaluation.}
KASA provides temporally consistent articulated supervision obtained from human-in-the-loop semantic annotation and differentiable pose tracking. During training, KVLR uses this supervision to construct the Kinematic-to-Visual Action Field as a conditioning input. During evaluation, however, the generated videos are first processed by an external visual extractor to obtain instrument masks or trajectories, and these extracted visual measurements are then compared with the projected action references. Therefore, the evaluation does not directly access the internal routed action features or the model's conditioning branch. Nevertheless, because the default evaluator uses the same semantic categories as KASA, we further introduce representation-independent checks below.

\paragraph{Cross-evaluator action-faithfulness evaluation.}
To reduce dependence on the annotation model used during KASA construction, we evaluate generated videos with an independent instrument extractor that is not used for constructing KASA or training KVLR. Specifically, we use a separately trained surgical instrument segmentation model (Grounded SAM~\cite{ren2024grounded}), and recompute Chamfer Distance (CD), Temporal IoU (TI), and Area Flicker (AF) on the validation set. As shown in Tab.~\ref{tab:cross_evaluator}, KVLR remains consistently better than the strongest action-conditioned baseline under both evaluators. This suggests that the improvement is not specific to the evaluation pipeline.

\begin{table}[t]
\centering
\caption{Cross-evaluator validation of action faithfulness. Grounded SAM~\cite{ren2024grounded} is independently trained and is not used for KASA construction or KVLR training. Results are averaged over the three KASA action subsets.}
\label{tab:cross_evaluator}
\renewcommand{\arraystretch}{1.08}
\setlength{\tabcolsep}{24pt}
\resizebox{\textwidth}{!}{
\begin{tabular}{llccc}
\toprule
Evaluator & Method & CD $\downarrow$ & TI $\uparrow$ & AF $\downarrow$ \\
\midrule
Grounded SAM~\cite{ren2024grounded} & Cosmos-H-Surgical$^{*}$~\cite{he2026cosmosh} & 94.65 & 0.78 & 8.52 \\
Grounded SAM~\cite{ren2024grounded}  & KVLR & {86.90} & {0.89} & {4.48} \\
\midrule
SAM2~\cite{ravi2024sam} & Cosmos-H-Surgical$^{*}$~\cite{he2026cosmosh} & 101.34 & 0.75 & 9.17 \\
SAM2~\cite{ravi2024sam} & KVLR & {90.82} & {0.86} & {5.03} \\
\bottomrule
\end{tabular}}
\end{table}

\paragraph{Trajectory-level verification independent of mask overlap.}
Mask-overlap metrics may still favor methods that better match the projected action masks. We therefore evaluate action consistency using trajectory-level measurements extracted from generated videos. For each generated clip, we estimate the visible tool centerline and tip trajectory using Grounded SAM~\cite{ren2024grounded}, and compare them with the projected reference trajectory. We report mean centerline error, endpoint error, and motion-direction accuracy. As shown in Tab.~\ref{tab:trajectory_evaluator}, KVLR also improves trajectory-level alignment, indicating that the generated motion better follows the specified articulated action beyond mask-overlap agreement.

\begin{table}[t]
\centering
\caption{Trajectory-level action verification with an independent visual tracker. Lower centerline and endpoint errors are better; higher direction accuracy is better.}
\label{tab:trajectory_evaluator}
\renewcommand{\arraystretch}{1.08}
\setlength{\tabcolsep}{15pt}
\resizebox{\textwidth}{!}{
\begin{tabular}{lccc}
\toprule
Method & Centerline Error (px) $\downarrow$ & Endpoint Error (px) $\downarrow$ & Direction Acc. (\%) $\uparrow$ \\
\midrule
Cosmos-H-Surgical$^{*}$~\cite{he2026cosmosh} & 11.83 & 15.62 & 71.4 \\
KVLR-fast & 8.05 & 10.24 & 82.1 \\
KVLR & \textbf{7.21} & \textbf{9.13} & \textbf{84.8} \\
\bottomrule
\end{tabular}
}
\end{table}

\paragraph{Action perturbation sanity check.}
Finally, we verify that KVLR uses the provided action trajectory rather than only relying on the reference image or dataset prior. We evaluate three perturbed-control settings: \textit{shuffled action}, where the action trajectory is randomly sampled from another clip of the same category; \textit{wrong-category action}, where the action is sampled from a different sub-action category; and \textit{reversed action}, where the temporal order of the action sequence is reversed. Tab.~\ref{tab:action_perturbation} shows that perturbing the action substantially degrades action-faithfulness metrics while only moderately affecting image-level fidelity. This confirms that the action input has a causal influence on the generated motion, and that the default improvements are not caused by simply replaying appearance priors from the reference image.

\begin{table}[t]
\centering
\caption{Action perturbation sanity check for KVLR. Perturbing the input action reduces action faithfulness, showing that the model depends on the specified action trajectory. Results are averaged over the validation subset.}
\label{tab:action_perturbation}
\renewcommand{\arraystretch}{1.08}
\resizebox{\textwidth}{!}{
\setlength{\tabcolsep}{30pt}
\begin{tabular}{lcccc}
\toprule
Control Input & CD $\downarrow$ & TI $\uparrow$ & AF $\downarrow$ & FID $\downarrow$ \\
\midrule
Matched action & \textbf{86.90} & \textbf{0.89} & \textbf{4.48} & \textbf{14.38} \\
Shuffled action & 126.75 & 0.64 & 10.82 & 18.94 \\
Wrong-category action & 138.42 & 0.58 & 12.36 & 21.15 \\
Reversed action & 119.63 & 0.67 & 9.74 & 17.88 \\
\bottomrule
\end{tabular}
}
\end{table}

\begin{figure}[!t]
\centering
\includegraphics[width=0.99\linewidth]{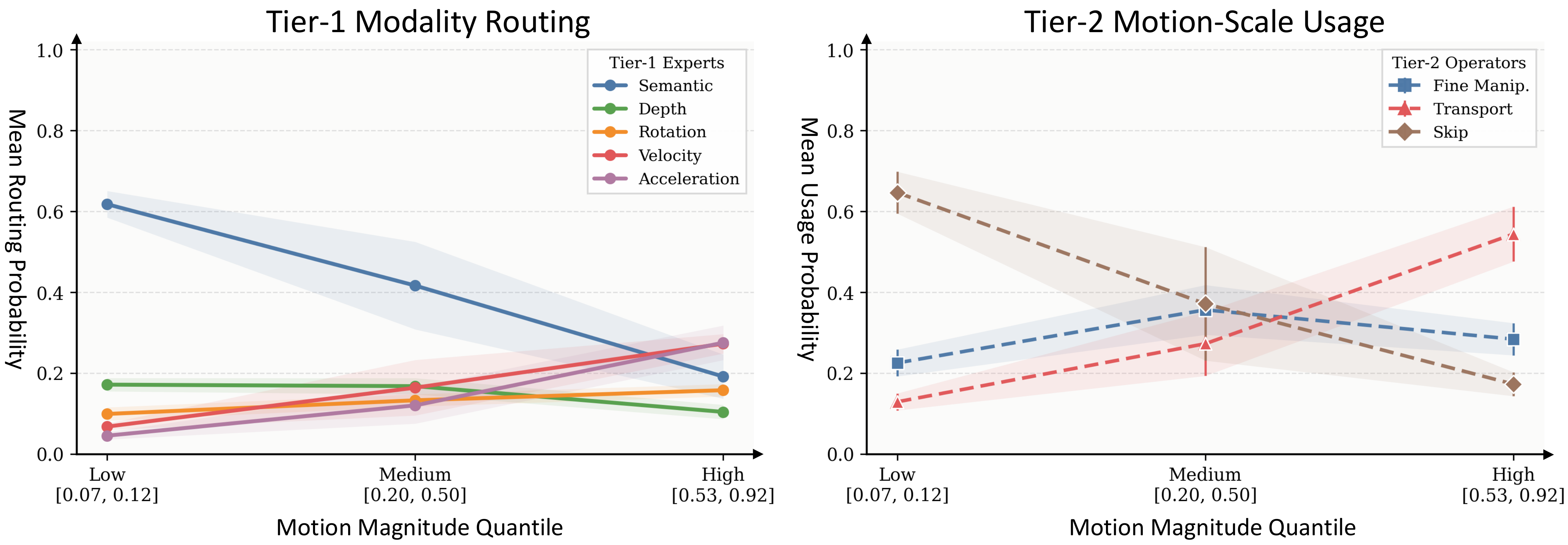}
    \caption{{Motion-specialized routing statistics.}
    We group tokens by motion-magnitude quantiles and report the mean routing mass with 95\% confidence intervals.
    At {Tier 1}, expert usage shifts from semantic-dominant routing in low-motion regions to velocity- and acceleration-dominant routing as motion increases, indicating that the model allocates more computation to dynamic cues when articulated changes become larger.
    At {Tier 2}, the routing pattern transitions from skip-dominant behavior in low-motion bins to transport-dominant behavior in high-motion bins, while fine-manipulation operators are most active in the intermediate regime.
    These results quantitatively confirm that the proposed hierarchical routing learns a motion-dependent decomposition of surgical control rather than a static or uniform expert allocation.}
 
    \label{fig:motion_specialization}
\end{figure}

\subsection{Quantitative Analysis of Hierarchical Expert Specialization}

Fig.~\ref{fig:motion_specialization} further quantifies how the learned routing behavior evolves with motion intensity.
We compute clip-level routing statistics over motion-magnitude quantiles and report the corresponding means and 95\% confidence intervals.
At Tier~1, the activation mass gradually shifts away from the semantic branch and toward the velocity and acceleration branches as motion magnitude increases, showing that the router does not amplify all experts uniformly, but selectively emphasizes dynamics-sensitive pathways when stronger articulated motion is present.
At Tier~2, the operator distribution changes from skip-dominant behavior in low-motion regions to transport-dominant behavior in high-motion regions, while fine-manipulation experts remain most competitive in the intermediate range.
This intermediate regime also exhibits wider confidence intervals, suggesting that it contains the most heterogeneous micro-dynamics across clips, where subtle local adjustments and larger motion transitions co-exist.
Overall, these statistics provide quantitative evidence that hierarchical routing captures a meaningful motion-specialized control decomposition, aligning low-motion states with semantic/static processing and high-motion states with transport-oriented dynamic computation.

\subsection{Detailed Efficient-Generation Ablation Breakdown}
\label{sec:app_eff_breakdown}

We expand the main efficiency ablation to isolate what each component contributes. For every row, we report the latency and FLOP speedup relative to the 50-step teacher, the marginal compute saving relative to the immediately preceding row, the CD-based control retention, and the FID increase relative to the teacher. The control-retention ratio is computed as $\mathrm{CD}_{\mathrm{teacher}}/\mathrm{CD}_{\mathrm{method}}$, since lower CD indicates better action alignment.

\noindent\textbf{Overall Interpretation.}
The baseline table separates three effects that are conflated in the final KVLR-fast number: few-step acceleration, action-aware preservation of the teacher's control policy, and routing-derived adaptive execution. Generic few-step distillation (DMD2)~\cite{yin2024improved} is computationally effective but action-destructive: it gives an $8.92\times$ latency speedup, yet retains only $88.3\%$ of the teacher's CD-based control accuracy and increases FID by $5.45$. The action-aware student keeps exactly the same measured cost as DMD2 but recovers the control retention to $99.3\%$ and reduces the FID increase to $+1.23$. Spatial adaptive execution and temporal caching then provide additional compute reductions, yielding the final $12.54\times$ latency speedup while retaining $96.7\%$ of teacher control accuracy.\looseness=-1

\noindent\textbf{Expanded Component Validations.}
To further isolate the internal dynamics of the efficient student and the adaptive execution policy, we introduce an expanded ablation matrix. Tab.~\ref{tab:expanded_eff_ablation} summarizes these diagnostic settings, evaluating the independent contributions of the distillation signals, spatial execution policies, and temporal refresh rates.

\begin{table}[!t]
\centering
\caption{Expanded Efficient-Generation Ablation. All metrics are evaluated on the KASA validation split. Latency and FLOPs are measured per 17-frame clip.}
\label{tab:expanded_eff_ablation}
\renewcommand{\arraystretch}{1.1}
\setlength{\tabcolsep}{15pt}
\resizebox{\linewidth}{!}{%
\begin{tabular}{llcccc}
\toprule
Group & Setting & \makecell{Latency\\(s) $\downarrow$} & \makecell{FLOPs\\(PF/clip) $\downarrow$} & \makecell{CD\\ $\downarrow$} & \makecell{FID\\ $\downarrow$} \\
\midrule
\textbf{Teacher} & Full 50-step structured generator & 4.64 & 6.94 & 86.89 & 14.37 \\
\midrule
\multirow{4}{*}{\makecell[l]{\textbf{Distillation Signal}\\ \textit{(Base: 4-step)}}} 
& Prediction-only (Generic) & 0.52 & 0.78 & 98.45 & 19.82 \\
& Prediction + Route & 0.52 & 0.78 & 92.10 & 17.50 \\
& Prediction + Control-path & 0.52 & 0.78 & 90.50 & 16.80 \\
& Pred + Route + Ctrl (Action-aware) & 0.52 & 0.78 & 87.50 & 15.60 \\
\midrule
\multirow{4}{*}{\makecell[l]{\textbf{Spatial Execution}\\ \textit{(Base: Action-aware)}}} 
& Full update for all tokens & 0.52 & 0.78 & 87.50 & 15.60 \\
& Full/Reuse only ($0.35, 0.00, 0.65$) & 0.38 & 0.57 & 91.20 & 17.90 \\
& Full/Light only ($0.20, 0.80, 0.00$) & 0.46 & 0.69 & 88.00 & 16.10 \\
& Full/Light/Reuse ($0.20, 0.30, 0.50$) & 0.42 & 0.63 & 88.65 & 16.55 \\
\midrule
\multirow{4}{*}{\makecell[l]{\textbf{Temporal Cache}\\ \textit{(Base: Spatial)}}} 
& Cache disabled ($1, 1, 1$) & 0.42 & 0.63 & 88.65 & 16.55 \\
& Conservative ($1, 2, 2$) & 0.39 & 0.59 & 89.10 & 17.05 \\
& Default intervals ($1, 2, 4$) & 0.37 & 0.55 & 89.87 & 17.67 \\
& Aggressive reuse ($1, 4, 8$) & 0.33 & 0.50 & 92.40 & 19.30 \\
\bottomrule
\end{tabular}%
}
\end{table}

\noindent\textbf{Analysis of Distillation Signals.}
The distillation ablation isolates the value of the structured control pathway. While prediction-only distillation (DMD2)~\cite{yin2024improved}  severely degrades action faithfulness (CD $98.45$), adding routing alignment explicitly informs the student which physical modalities are active, dropping CD to $92.10$. However, preserving the intermediate action-control features (Control-path) provides an even stronger independent signal (CD $90.50$). The combination of both yields the full action-aware student (CD $87.50$), proving that route and control-path supervision are synergistic and transfer the teacher's action policy at zero additional inference cost.

\noindent\textbf{Analysis of Spatial Execution.}
The spatial execution ablation tests whether the scaled ``light'' branch is more effective than hard token suppression. A strict Full/Reuse policy (allocating $35\%$ of tokens to full updates and aggressively reusing the rest to match the target FLOPs) drops latency to $0.38$\,s but penalizes quality heavily (CD $91.20$), as critical edge cases are suppressed. Introducing the ``light'' branch (Full/Light/Reuse at $20/30/50$) allows the model to maintain broad contextual awareness with cheaper cached-routing updates, yielding a strictly better Pareto frontier (Latency $0.42$\,s, CD $88.65$).\looseness=-1

\noindent\textbf{Analysis of Temporal Cache.}
The temporal cache ablation exposes the deployment trade-offs of feature caching. Disabling the cache entirely relies strictly on spatial token sparsification ($0.42$\,s). Enabling conservative caching ($1,2,2$) yields ``free'' latency improvements ($0.39$\,s) with negligible CD degradation ($+0.45$). Our default ($1,2,4$) setting pushes this efficiency further. However, aggressive cache reuse ($1,4,8$) introduces noticeable temporal drift; as the instrument moves out of the cached bounding regions, the student fails to update the active kinematics, resulting in a sharp penalty to both CD ($92.40$) and visual fidelity (FID $19.30$).

\begin{figure}[!t]
    \centering
    \includegraphics[width=1.0\linewidth]{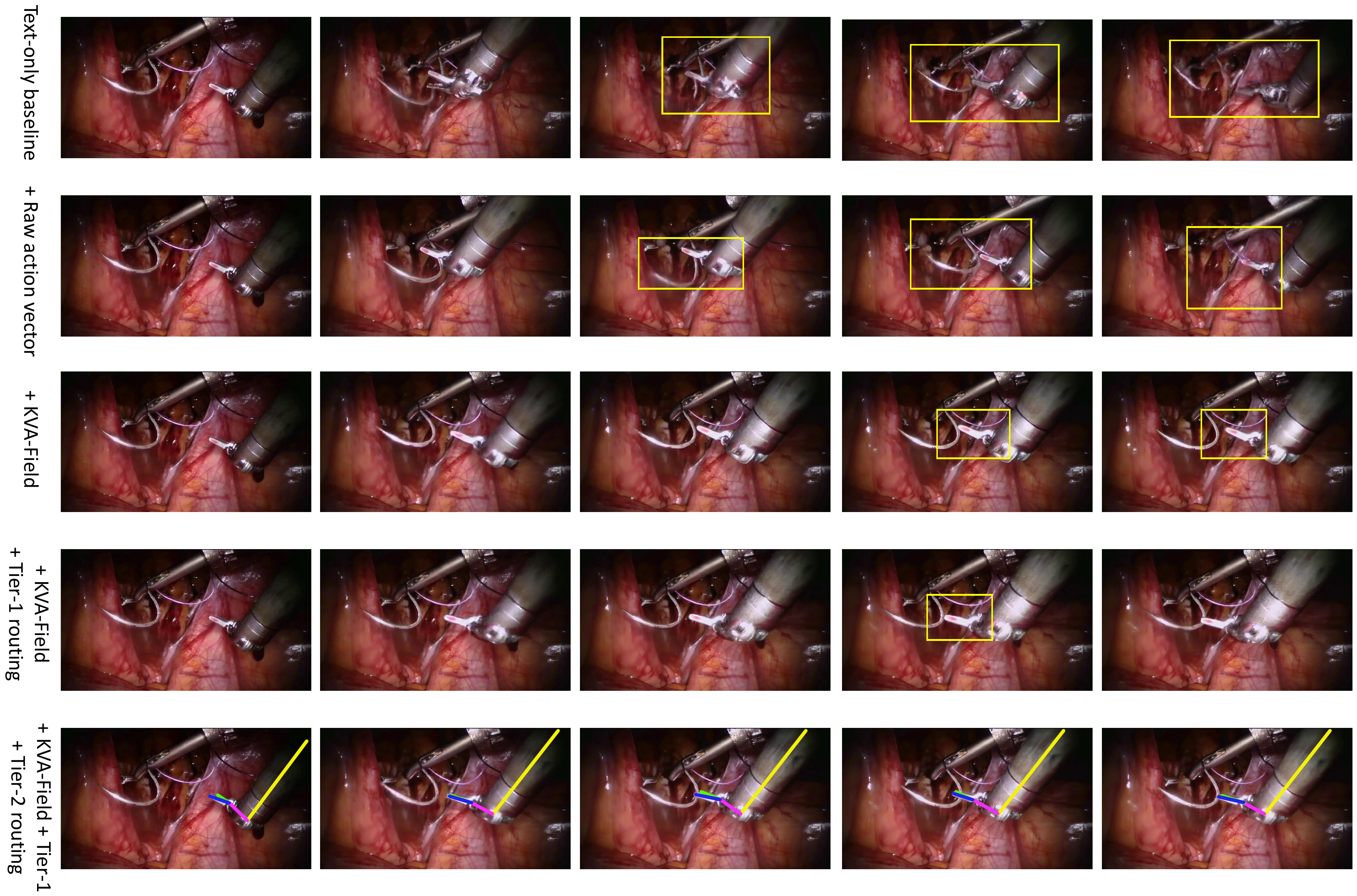}
    \caption{{Additional qualitative comparisons on architecture design.}
    We compare representative variants from the architecture ablation, including text-only generation,
    direct raw-action conditioning, KVA-Field conditioning, Tier-1 routing, and the full hierarchical
    routing model. The full model produces clearer tool appearance and more action-consistent temporal
    evolution.}
    \label{supple_abl_arc}
\end{figure}

\begin{figure}[!t]
    \centering
    \includegraphics[width=1.0\linewidth]{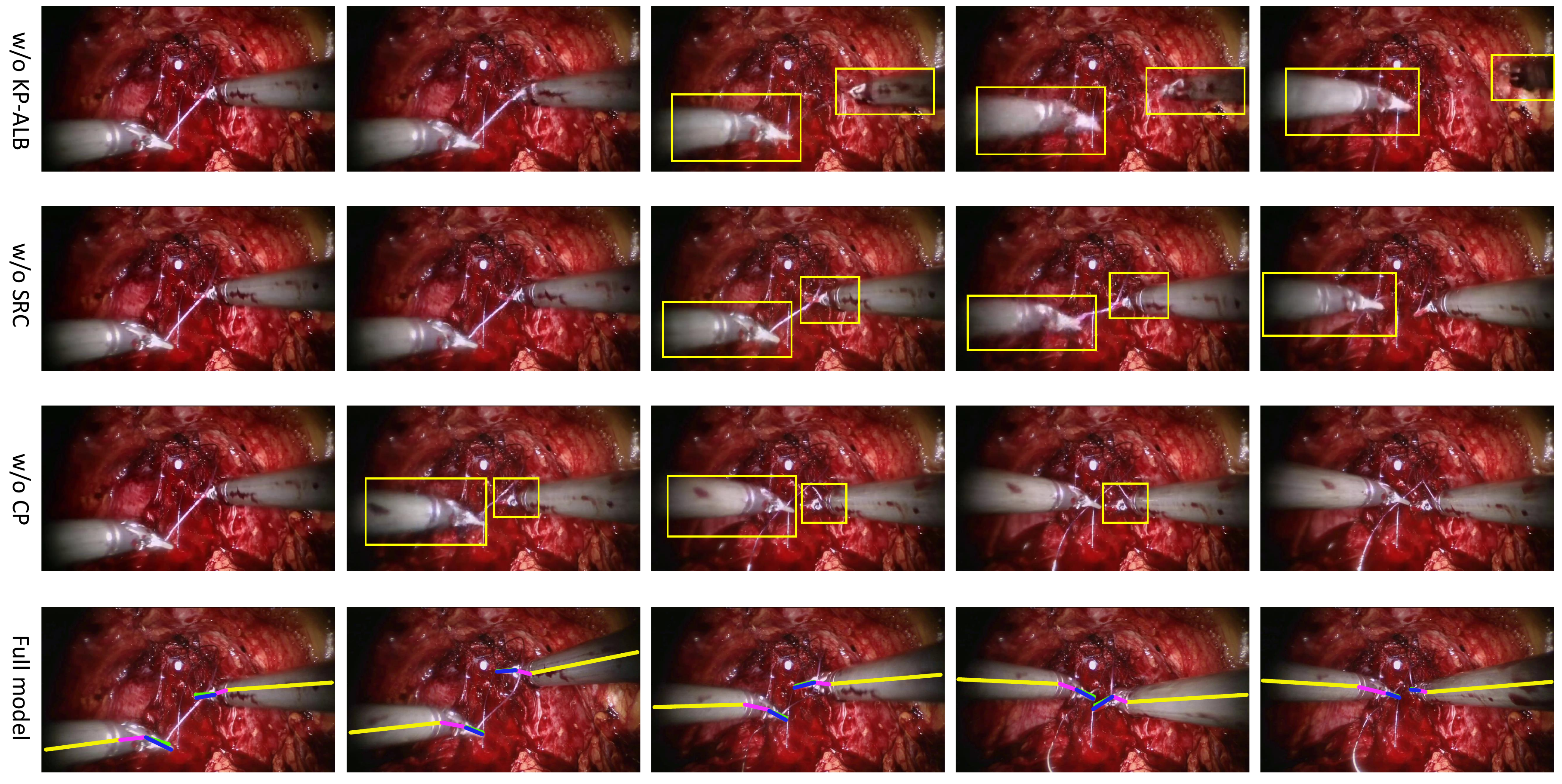}
    \caption{{Additional qualitative comparisons on kinematic-prior losses.}
    We compare the full model with variants that remove KP-ALB, SRC, or CP. The proposed
    kinematic-prior losses improve spatial alignment and temporal consistency, leading to more stable action-conditioned generation.\looseness=-1}
    \label{supple_abl_loss}
\end{figure}

\subsection{More Qualitative Results.}\label{appendix:qua}

\noindent\textbf{Qualitative results on architecture design.}
Fig.~\ref{supple_abl_arc} provides additional qualitative comparisons for the architecture ablation.
The text-only baseline fails to maintain a clear instrument appearance and exhibits inaccurate tool motion, especially in the later frames. Directly injecting raw action vectors yields only limited improvement: although the generated motion is partially guided by the input actions, the instrument trajectory remains imprecise, and the visual quality degrades over time. This supports our observation that low-dimensional articulated kinematics are difficult to use directly for dense image-space video evolution. Replacing raw actions with the proposed KVA-Field substantially improves the spatial alignment between action inputs and visual generation, leading to clearer tool structures and more consistent motion. Introducing Tier-1 routing further improves the use of different physical control modalities, while the full model with both Tier-1 and Tier-2 routing achieves the best qualitative results by adapting control to both modality types and motion scales. These qualitative trends are consistent with the quantitative architecture ablation in Tab.~\ref{tab:ablation_core}, showing that action lifting and hierarchical routing are jointly important for action-controlled surgical video generation.\looseness=-1

\begin{figure}[!t]
    \centering
    \vspace{-0pt}
    \includegraphics[width=1.0\linewidth]{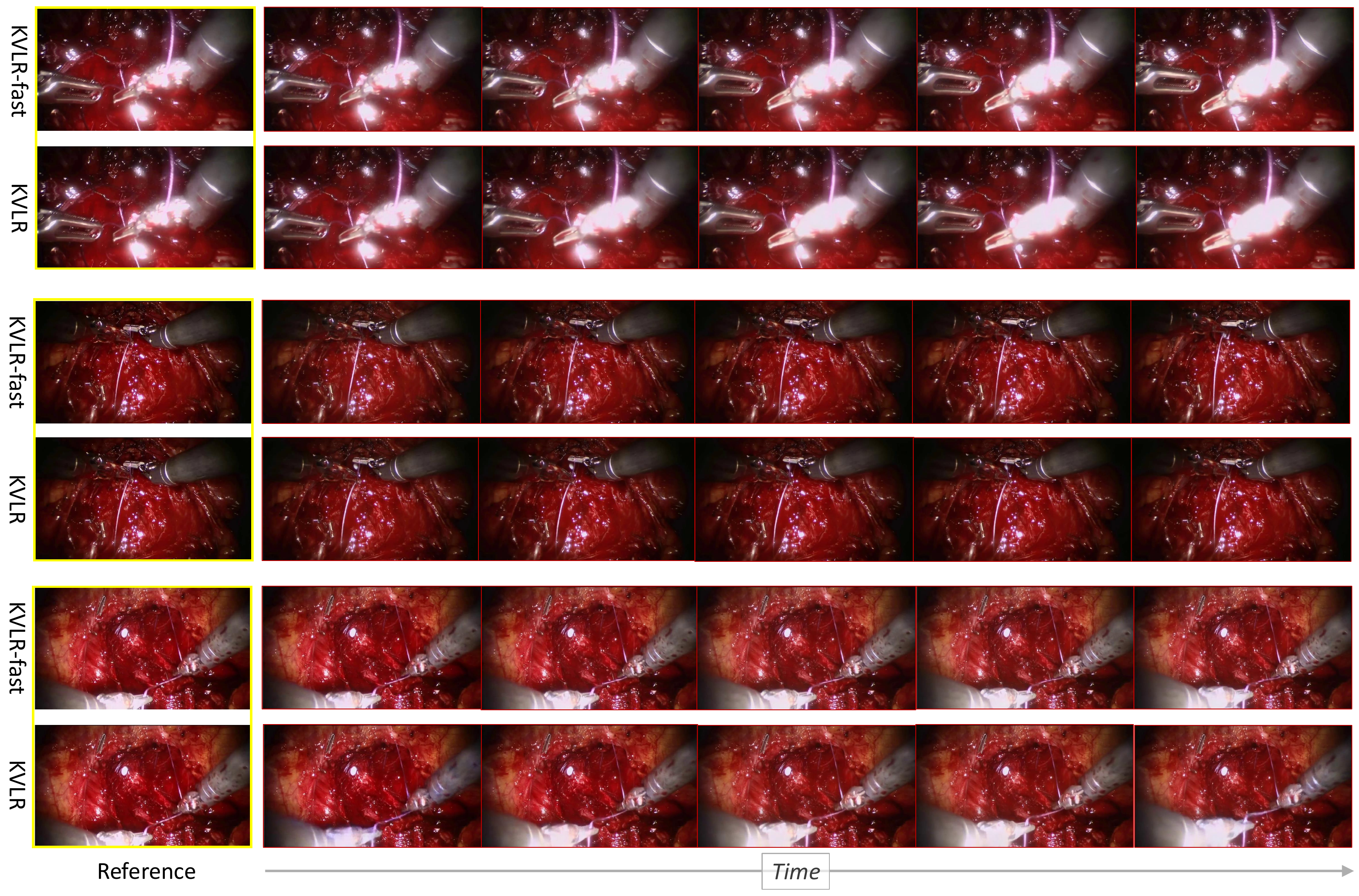}
    \caption{{Additional generation results of KVLR and KVLR-fast.}
    We show more examples across diverse surgical scenes and actions.
    Both the full model and the efficient variant preserve action-consistent visual evolution, while KVLR generally yields slightly sharper appearance details and stronger temporal stability under complex motion.
    These examples further show that the action-adaptive efficient model retains most of the control fidelity of the full generator despite using substantially reduced computation.}
    \label{fig_more_diverse}
\end{figure}

\noindent\textbf{Qualitative results on kinematic-prior losses.}
Fig.~\ref{supple_abl_loss} shows additional qualitative comparisons for the kinematic-prior loss ablation. Replacing KP-ALB with uniform load balancing~\cite{ruby2020binary,fei2024scaling,zheng2025dense2moe} weakens the correspondence between routing decisions and physical action cues, resulting in degraded
spatial alignment and blurrier tool appearance. Removing SRC leads to visibly less stable temporal evolution, with tool structures and interaction regions becoming inconsistent in the later frames.
This suggests that temporal regularization of routing decisions is important for maintaining coherent action-conditioned generation across video frames. Removing CP also degrades the results, indicating that predictable routing helps stabilize expert selection and supports reliable control during generation.
Overall, the full model produces the most stable and visually faithful results, confirming that the three kinematic-prior losses are complementary and collectively improve the quality of routed visual control.\looseness=-1

\begin{figure}[!t]
    \centering
    \vspace{-0pt}
    \includegraphics[width=1.0\linewidth]{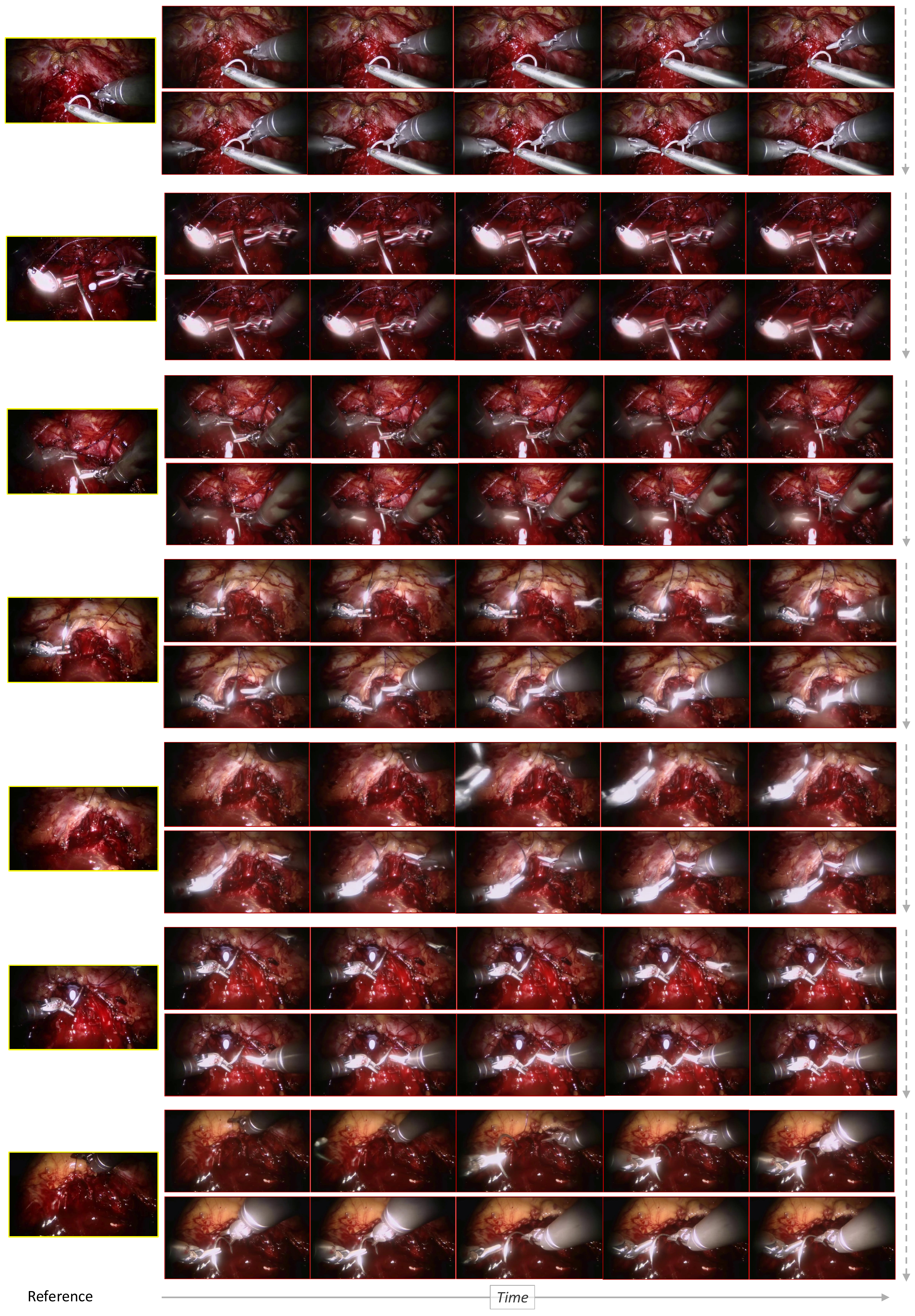}
    \vspace{-0pt}
    \caption{{Additional diverse generation results of KVLR on different surgical actions.}
    The generated videos remain consistent with the prescribed articulated controls across different motion patterns, including fine manipulation, larger transport motion, and varying tool--tissue interaction stages.
    These results highlight the robustness of the proposed structured control interface across multiple action categories and scene configurations.}
    \label{fig_more_diverse1}
    \vspace{-0pt}
\end{figure}

\noindent\textbf{Additional qualitative results.} We provide additional qualitative results in Figs.~\ref{fig_more_diverse} and ~\ref{fig_more_diverse1}.
These visualizations cover a wider range of scenes, action trajectories, and procedural stages across \textit{Knotting}, \textit{NeedleGrasping}, and \textit{NeedlePuncture}, and further illustrate two key properties of the proposed framework.
As shown in Fig.~\ref{fig_more_diverse}, both KVLR and KVLR-fast preserve action-consistent scene evolution under diverse articulated controls, including large tool transport, fine local adjustment, and contact-heavy interaction.
Moreover, as illustrated in Figs.~\ref{fig_more_diverse1}, the generated videos remain visually coherent across challenging conditions such as appearance variation, illumination change, cluttered backgrounds, and different instrument poses.
Together with the main qualitative comparisons, these results further support that the proposed structured kinematic-to-visual control improves both controllability and visual stability across heterogeneous surgical dynamics.

\end{document}